\documentclass{article}

\usepackage[nonatbib, final]{nips_2017_arxiv}

\usepackage[utf8]{inputenc} % allow utf-8 input
\usepackage[T1]{fontenc}    % use 8-bit T1 fonts
\usepackage{hyperref}       % hyperlinks
\usepackage{url}            % simple URL typesetting
\usepackage{booktabs}       % professional-quality tables
\usepackage{amsfonts}       % blackboard math symbols
\usepackage{nicefrac}       % compact symbols for 1/2, etc.
\usepackage{microtype}      % microtypography

\usepackage{amsmath,amsfonts,amssymb,amsbsy,amsmath}
\usepackage[pdftex]{graphicx}
\usepackage{subfig}
\usepackage{caption}
\usepackage{algorithm,algorithmic}
\usepackage{tikz}
\newcommand*\circled[1]{\scriptsize \tikz[baseline=(char.base)]{
            \node[shape=circle,draw,inner sep=2pt] (char) {#1};}}

\newtheorem{defn}{Definition}
\newtheorem{thm}{Theorem}

\title{Convolutional Networks with MuxOut Layers as Multi--rate Systems for Image Upscaling}

\author{
  Pablo Navarrete~Michelini \\
  BOE Technology Group Co., Ltd.\\
  Beijing, China \\
  \texttt{pnavarre@boe.com.cn} \\
  \And
  Hanwen Liu \\
  BOE Technology Group Co., Ltd.\\
  Beijing, China \\
  \texttt{lhw@boe.com.cn} \\
}

\begin{document}

\maketitle

\begin{abstract}
  We interpret convolutional networks as adaptive filters and combine them with so--called \emph{MuxOut} layers to efficiently upscale low resolution images. We formalize this interpretation by deriving a linear and space--variant structure of a convolutional network when its activations are fixed. We introduce general purpose algorithms to analyze a network and show its overall filter effect for each given location. We use this analysis to evaluate two types of image upscalers: deterministic upscalers that target the recovery of details from original content; and second, a new generation of upscalers that can sample the distribution of upscale \emph{aliases} (images that share the same downscale version) that look like real content.
\end{abstract}

\section{Introduction}
Image upscaling has been studied for decades and remains an active topic of research because of constant technological advances in digital imaging. One scenario where image upscaling is now more demanding arises in digital television with the introduction of new standards \cite{MSugawara_2014a}. Large upscaling factors are needed to convert standard--definition resolutions to the upcoming 8K UHD resolutions.

Let $\mathcal{H}$ be the set of all high--resolution images of size $H\times W$, let $\mathcal{H}^R\subset\mathcal{H}$ be the subset of high--resolution images that correspond to a particular real environment (e.g. photographs, or drawings, or text, etc.), and $\mathcal{L}$ the set of all low--resolution images of size $h\times w$ (with $h<H$, $w<W$). We are interested in the set of \emph{aliased} images:
\begin{equation}
    \mathcal{A} = \left\{ X,Y \in \mathcal{H} \quad s.t. \quad S_{down}(X)=S_{down}(Y) \right\} \;,
\end{equation}
where $S_{down}:\mathcal{H}\rightarrow\mathcal{L}$ is a \emph{down--scale} operator. Particularly, we are interested in the set $\mathcal{A}^R=\mathcal{A}\cap\mathcal{H}^R$ of alias images that correspond to real content. An upscaler aims to find, or approximate, an element $X$ of $\emph{A}^R$ such that $S_{down}(X)$ is equal to a given low--resolution image $x$. We expect this to be possible because our image $x$ has been obtained by a procedure equal or similar to $S_{down}$ (e.g. software downscaler, sensor arrays, etc.).

Upscaling signals by integer factors is understood in classical interpolation theory as two sequential processes: upsample (insert zeros) and filter \cite{PJDavis_1976a, JGProakis_2007a}. In order to avoid multiplying by zeros, a more efficient and equivalent process is to: split the filter into a bank of filters and multiplex the outputs \cite{PPVaidyanathan_1993a, JGProakis_2007a, SMallat_1998a}. Standard upscaler algorithms, such as bicubic or Lanczos, do not have references of $\mathcal{A}^R$ and choose to find high--resolution images with a narrow frequency content, using fixed low--pass filters. Similarly, more advanced upscalers (some called super--resolution) try to follow geometric principles to get closer to realistic images in $\mathcal{A}^R$. For example, \emph{edge--directed interpolation} uses adaptive filters to improve edge smoothness \cite{VRAlgazi_1991a, XLi_2001a}, or \emph{bandlet} methods use both adaptive upsampling and filtering \cite{SMallat_2007a}. More recently, machine learning has been able to use training samples $\mathcal{S}\in\mathcal{H}^R$ as a reference path to approach $\mathcal{A}^R$ \cite{SCPark_2003a}. In some cases, the optimization approach of machine learning hides the connection with classical interpolation theory (e.g. sparse representation with dictionaries \cite{JYang_2008a, JYang_2010a}). In other cases, the adaptive filter approach is explicit (e.g. RAISR \cite{DBLP:journals/corr/RomanoIM16}). Here, we will use the term \textbf{super--resolution} (SRes) for methods focused on interpolation, that aim to find one particular $X$ such that $x=S_{down}(X)$. We will call \textbf{hyper--resolution} (HRes) a more recent type of methods that aim to generate samples $\mathcal{G}\in\mathcal{A}^R$ such that $x=S_{down}(X), \forall X\in\mathcal{G}$. This approach has recently become feasible for applications by using \emph{generative adversarial networks} \cite{DBLP:journals/corr/LedigTHCATTWS16, AffGAN, PNavarrete_2017a, FaceSR}.

In this work, we focus on the problem of image upscaling, including SRes and HRes methods, for color images using convolutional networks. This line of work started with SRCNN \cite{CDong_2014a, CDong_2015a} motivated by the success of  deep--learning methods in image classification tasks \cite{YLeCun_2015a} and establishing a strong connection with sparse coding SRes methods \cite{JYang_2008a, JYang_2010a}. Our system shares strong connections with SRCNN but follows a different motivation. Namely, we aim to reveal a strong connection between convolutional--networks and multi--rate signal processing. By doing so, we can recover the classic interpretation of adaptive filters which can help us to design better architectures.  Thus, we aim to prove that convolutional networks are a natural and convenient choice for image upscaling tasks.

Regarding \textbf{system architecture}, we show how to embed the classical processes of filtering and muxing within the network. We improve a so-called \emph{MuxOut} layer \cite{PNavarrete_2016a} to work well with SGD--based algorithms and arbitrary datasets. Our solution proves to be stable and allows us to consider flexible architectures using multiple \emph{MuxOut} layers. We are able to configure a color image upscaler that balances lumminance versus color, similar to chroma--subsampling in image compression systems.

Regarding \textbf{network analysis}, we formalize our adaptive--filtering interpretation with novel techniques to visualize the structure of a network. Our analysis can display adaptive interpolation coefficients that allow comparisons with classical and advanced upscalers. The analysis shares the motivation of previous visualization techniques \cite{DBLP:conf/eccv/ZeilerF14} and can be used to analyze most deep--learning architectures.

\section{Upscaling with Convolutional Networks}
An \emph{upsampling} operator copies the pixel values of $x$, of size $h\times w$, into a zero image of size $M_x h \times M_x w$, placing values in a regular grid. There are $M=M_x\cdot M_y$ possible distinct grids where the values of $x$ can be copied. Thus, we define $M$ upsampling operators as follows:
\begin{defn}[Up/Down--sampling Operators]
    Let $x\in\mathbb{R}^{h\times w}$ be an image of size $h\times w$. The up--sampling operators $U^n$ with $n=0,\ldots,M-1$ is given by:
    \begin{equation}
        \left( U^n x_{i,j} \right)[p,q] = \begin{cases}
            x_{i,j} & \quad\text{if } p=M_y i+a_n \; \wedge \; q=M_x j+b_n,\\
            & \quad a_n=(n-1) \text{mod } M_y, \quad b_n = \lfloor (n-1)/M_y \rfloor, \\
            0 & \quad\text{otherwise,}
        \end{cases}
    \end{equation}
    where $i=0,\ldots,h-1$, $j=0,\ldots w-1$, $p=0,\ldots,M_y h-1$ and $q=0,\ldots,M_x w-1$.

    Correspondingly, we called down--sampling the transposed operators $D^n$ given by:
    \begin{equation}
        \left( D^n x_{p,q} \right)[i,j] = x_{M_y i+a_n, M_x j+b_n} \;.
    \end{equation}
\end{defn}

Our purpose now is to embed the classical processes of muxing within the network. Initial efforts in this direction using a \emph{MuxOut} layer worked well only with simple and particular datasets \cite{PNavarrete_2016a}. Similar efforts, using a \emph{sub--pixel convolutional network} have succeded by restricting all the muxing process to one layer \cite{DBLP:journals/corr/ShiCHTABRW16}. We identify two conflicting problems when using \emph{MuxOut} within a network: upsampling and filtering. One problem is combinatorial (where to place pixels in larger images?) and the other is an interpolation problem (how to weight neighboring pixels?). Filter weights in convolutional networks can potentially solve both problems by shifting and adjusting weights, but empirical tests show visible artifacts or unstable training with SGD--based methods \cite{odena2016deconvolution, PNavarrete_2016a}. We propose an improved \emph{MuxOut} layer that solves the combinatorial problem by explicitly giving the network different muxing combinations. The network can now adaptively select which combination works better using activation layers, and filter weights can now focus on the interpolation problem.

\begin{defn}[MuxOut/T--MuxOut Layers]
    Let $x\in\mathbb{R}^{G_\text{in}M\times h\times w}$ be $G_\text{in}M$ feature images of size $h\times w$. Let $x^g\in\mathbb{R}^{G_\text{in}\times h\times w}$ be a group of $G_\text{in}$ input features such that $x=[x^1 \cdots x^M]^T$. A MuxOut layer maps $G_\text{in}$ groups of input features into $G_\text{out}$ groups of output features such that:
    \begin{equation}
        \text{MuxOut}_{M_x \times M_y}(x) = \left[\begin{matrix}
        U^{P_1(1)} & \cdots & U^{P_1(M)} \\
            & \ddots & \\
        U^{P_{G_\text{out}}(1)} & \cdots & U^{P_{G_\text{out}}(M)} \\
        \end{matrix}\right] \;
        \left[\begin{matrix}
        x^1 \\
        x^2 \\
        \vdots \\
        x^M \\
        \end{matrix}\right]
    \end{equation}
    where $P_k:\mathbb{N}\rightarrow\mathbb{N}, k=1,\ldots,G_\text{out}$ represent $G_\text{out}$ different permutations of the set $\left\{1,\ldots,G_\text{out}\right\}$.

    Correspondingly, we called T--MuxOut the transposed of a MuxOut operator given by:
    \begin{equation}
        \text{T--MuxOut}_{M_x \times M_y}(x) = \left[\begin{matrix}
        D^{P_1(1)} & \cdots & D^{P_{G_\text{out}}(1)} \\
            & \ddots & \\
        D^{P_1(M)} & \cdots & D^{P_{G_\text{out}}(M)} \\
        \end{matrix}\right] \;
        \left[\begin{matrix}
        x^1 \\
        x^2 \\
        \vdots \\
        x^M \\
        \end{matrix}\right] \;.
    \end{equation}
\end{defn}

\begin{figure}[t]
  \centering
  \subfloat[MuxOut $2\times 2$ layer.]{\includegraphics[width=0.3\linewidth]{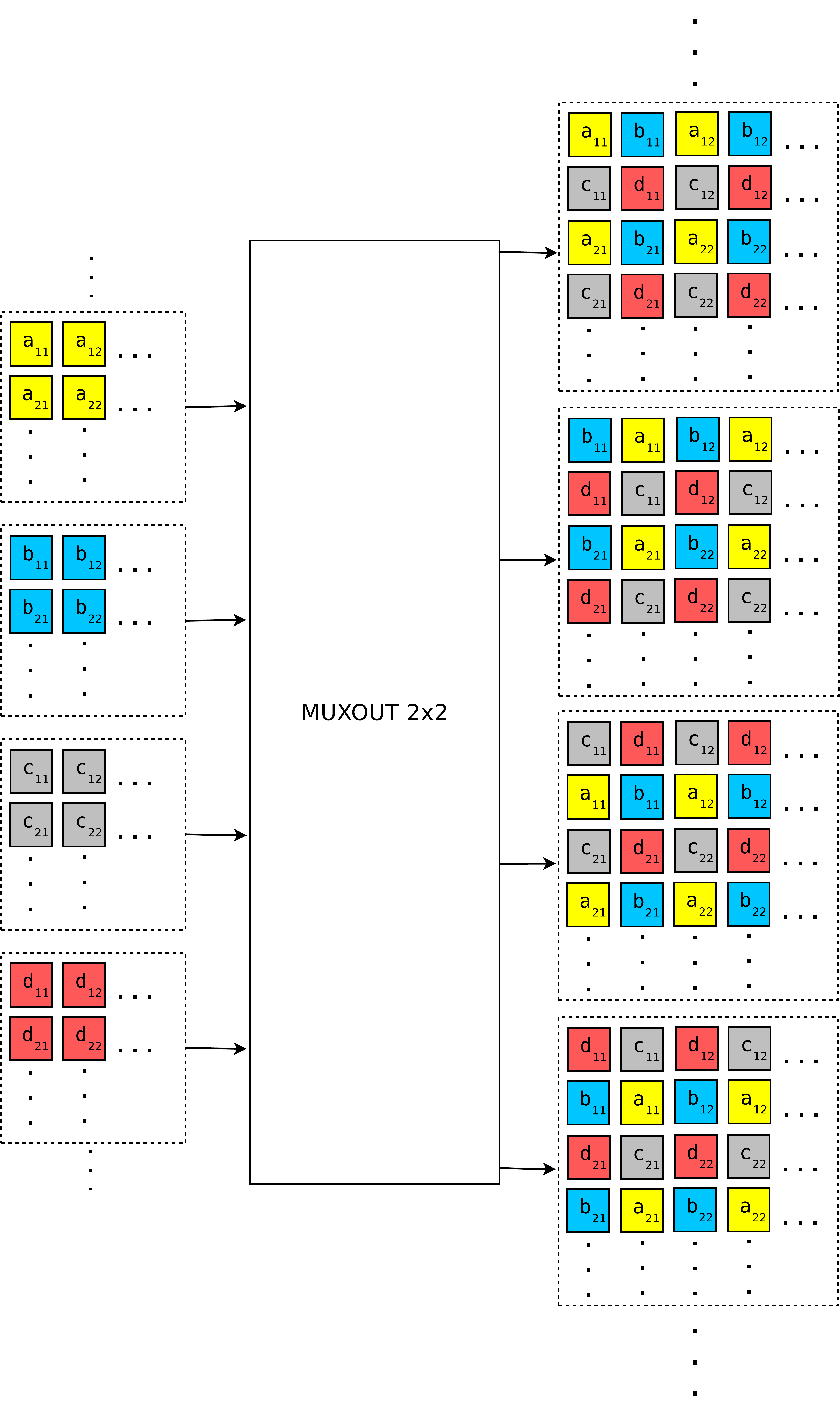} \label{fig:muxout}} \hfil
  \subfloat[Transposed MuxOut $2\times 2$ layer]{\raisebox{0.03\linewidth}{\includegraphics[width=0.32\linewidth]{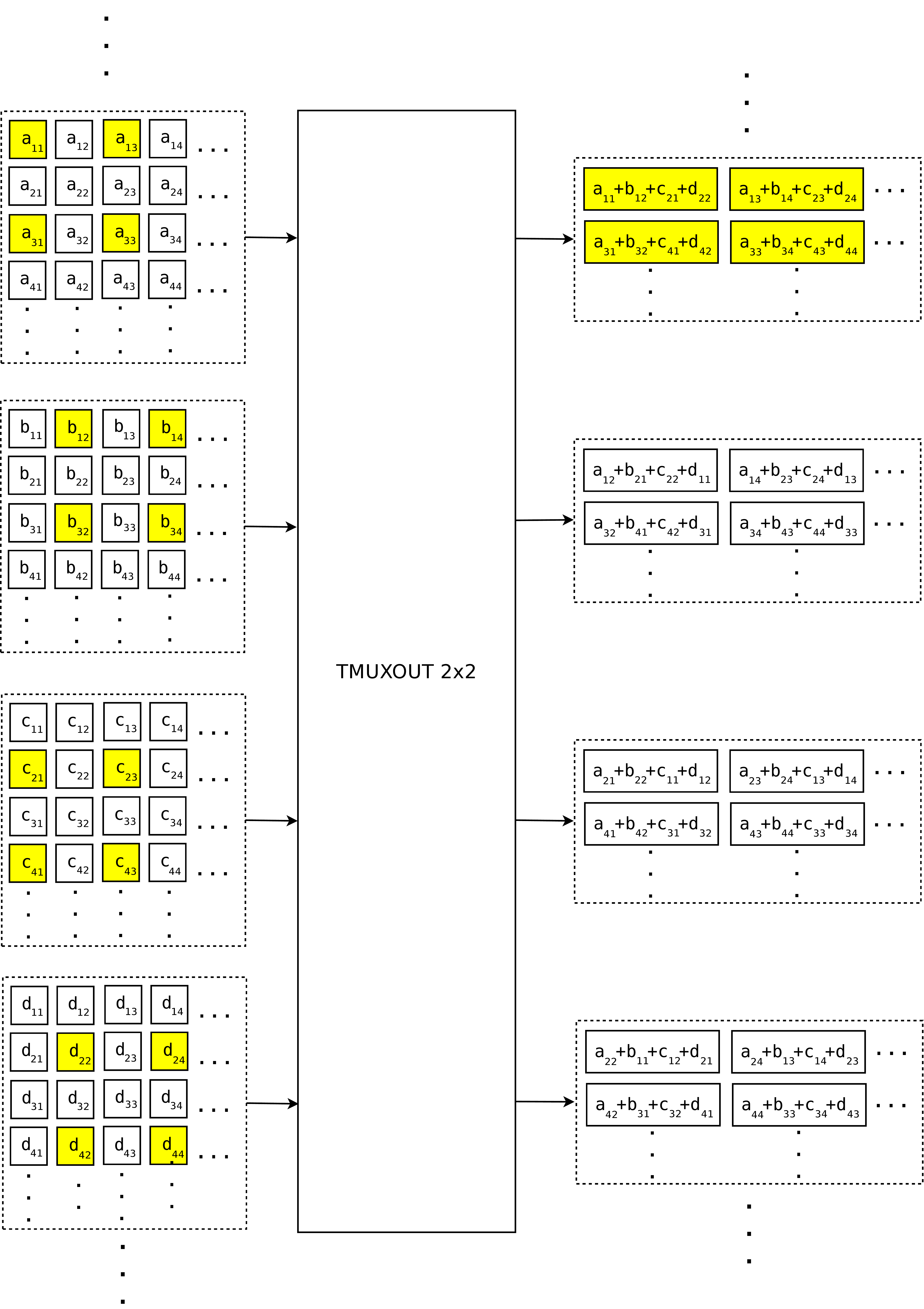}} \label{fig:tmuxout}}
  \raisebox{0.15\linewidth}{\begin{tabular}{c}
      \subfloat[Muxout recommended configuration using definitions \ref{eq:conv_activ}.]{\includegraphics[width=0.3\linewidth]{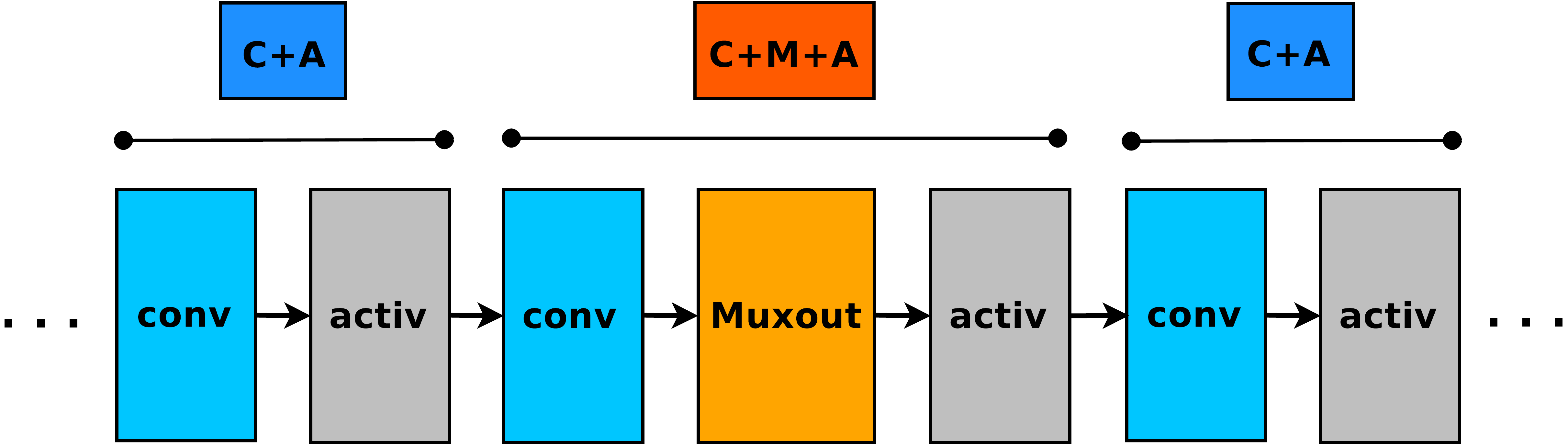} \label{fig:muxout_use}} \vspace*{0.1\linewidth} \\
      \subfloat[T--MuxOut recommended configuration follows standard use of pooling layers.]{\includegraphics[width=0.3\linewidth]{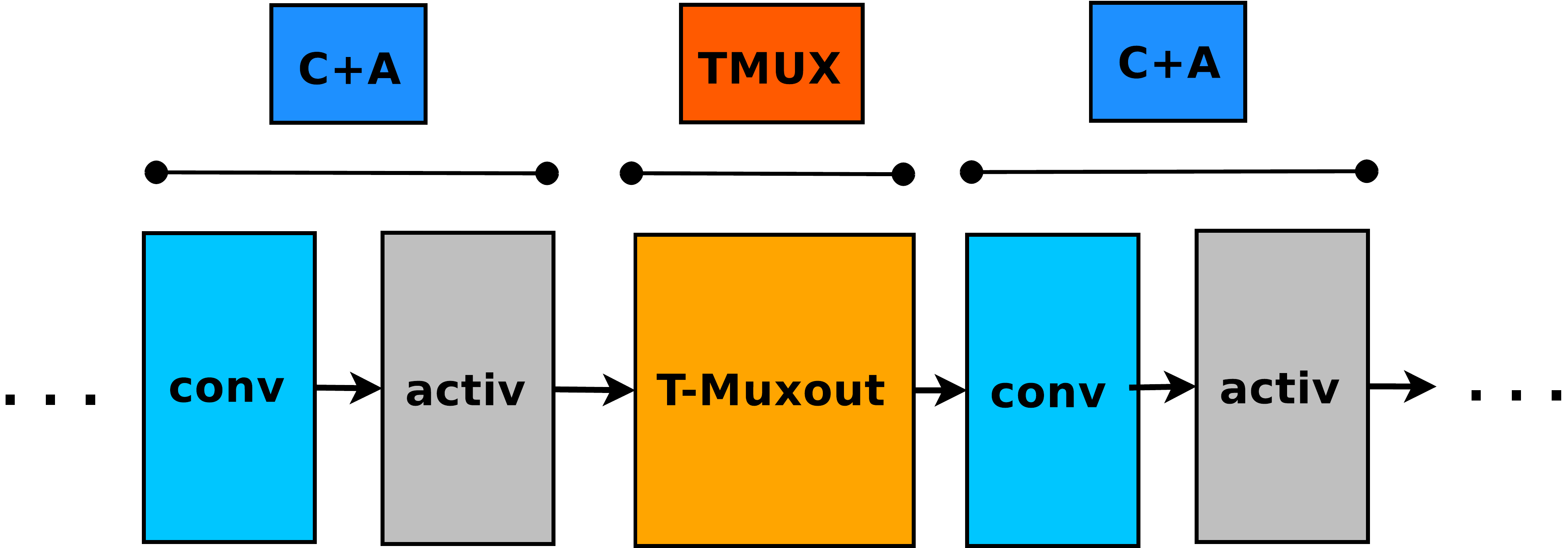} \label{fig:tmuxout_use}} \vspace*{0.1\linewidth}
  \end{tabular}} \vspace*{-0.07\linewidth}
  \caption{A \emph{MuxOut} $M_x\times M_y$ layer merges groups of $M_x\cdot M_y$ features like classic muxers in efficient upscalers. The layer and its transposed version consider different combination of pixels to avoid filter weights to solve a combinatorial problem during training.}
\end{figure}

Of particular interest is the case of circular permutations where $G_\text{in}=G_\text{out}$, which we will use in our experiments. We can recover the original MuxOut layer in \cite{PNavarrete_2016a} by using $G_\text{out}=1$. MuxOut and T--Muxout require the input number of features to be a multiple of $M = M_x \cdot M_x$, and they can be used anywhere within a convolutional network as any other pooling or unpooling layer. Examples of MuxOut and T--Muxout layers for $M_x=M_y=2$ are shown in Figure \ref{fig:muxout} and \ref{fig:tmuxout}, respectively.

It is common to define a convolutional layer including convolutions, biases and activation functions all together. Here, we will consider a convolutional transformation (the filtering part) and activation with bias (the switching part) as independent layers. This is,
\begin{equation}
    \text{conv}(x^{f_{in}}) = \sum_{f_{in}} x^{f_{in}} \ast w^{f_{in}, f_{out}}\;, \quad
    \text{activ}(x^f) = \sigma\left(x^f + b^f\right)  \label{eq:conv_activ}
\end{equation}
where $f$ represents a feature number, $\ast$ is a convolution operator, and $w^{f_{in}, f_{out}}$ is a given filter connecting features $f_{in}$ and $f_{out}$. Using this definition we then recommend to use MuxOut layers between convolution and activation layers as shown in Figure \ref{fig:muxout_use}. We propose to apply activation after muxing operations in order to force biases to activate/deactivate features already combined.

\section{Semi--classical Network Analysis}
\label{analysis}

A convolutional network is a sequence of linear transformations and nonlinear activations. This is,
\begin{eqnarray}
    y_n & = & W_n x_{n-1} + b_n \label{eq:linear} \\
    x_n & = & \sigma\left(y_n\right) \label{eq:activation} \;,
\end{eqnarray}
where $x_n$ and $y_n$ are vectorized versions of feature vectors at layer $n$, after and before activations, respectively. The parameters of the network at layer $n$ are contained in the biases $b_n$ and the sparse matrices $W_n$ representing the linear transformation of the convolutional operators. The input of the network is the vector $x_0$ and the output, after $n$ convolutional layers, is $x_n$.

\begin{defn}[Activation Mask]
    The \emph{mask} of an activation function $\sigma$ is given by:
    \begin{equation}
        \Sigma(x)[i] = \begin{cases}
            \sigma(x[i])/x[i] & \quad\text{if } x[i]\neq 0\\
            1 & \quad\text{otherwise}
        \end{cases}
    \end{equation}
\end{defn}
The \emph{mask} of an activation $\sigma$ measures the component--wise gain of the activation layer. For ReLU, the mask only takes values $0$ or $1$.

\begin{thm}[Fixed Activation Structure] \label{thm:fixed_act}
Let $\hat{W}_n = \Sigma(y_n) W_n$ and $\hat{b}_n = \Sigma(y_n) b_n$ be the \emph{masked filters} and \emph{biases} of a network for a given input $x_n$. Let
\begin{equation}
    F_0 = I \;, \quad F_i = \prod_{k=n-i+1}^n \hat{W}_k, \quad\text{for } i=1,\ldots,n \;,
\end{equation}
be the \emph{accumulated masked filters} at layer $i$. The output of the convolutional network is given by:
\begin{equation}
    x_n = W_\text{eff} \; x_0 + b_\text{eff} \;,
\end{equation}
where $W_\text{eff} = F_n$ is the \emph{effective filter} and $b_\text{eff} = F \ast \hat{b} = \sum_{k=0}^n F_k \hat{b}_{n-k}$ is the \emph{effective bias}.
\end{thm}
The proof of the theorem is a direct consequence of $\sigma(x) = \Sigma(x) x$ and expansion of \ref{eq:linear} and \ref{eq:activation}.

As a consequence, we can see the overall effect of a convolution network as a linear transformation where all non--linearities are contained in the effective filters/biases and are due exclusively to the gain of activation layers. In particular, if we fix the activations in the network, by recording the values of $\Sigma(y_n)$ for a given input, we can attempt to compute $W_{\text{eff}}$ and $b_{\text{eff}}$. After fixing the activations, $b_{\text{eff}}$ can be recovered with a zero input $x_0=0$, and $W_{\text{eff}}$ can be recovered using a set of inputs $X_0=I$, where $I$ is an identity matrix. This corresponds to a set of Kroneckder deltas $\delta_k$, also called \emph{impulse} vectors. In linear system terminology we are computing the \emph{impulse response} of the system \cite{JGProakis_2007a}. Linear layers such as pooling, unpooling, flattening, can also be included in the analysis. In case of max--pooling, the response must be recorded as done in previous visualization techniques \cite{DBLP:conf/eccv/ZeilerF14}.

The computation of all values of the sparse matrix $W_{\text{eff}}$ can be computationaly intensive and hardly necessary. For upscalers we are especially interested on the columns of $W_{\text{eff}}$ (sparse) that represent the adaptive interpolation filter for a given location. We can then compute the outputs of the system for impulses $\delta_k$ with $k$ in a subset of all possible columns $\mathcal{C}\subset\left\{1,\ldots,B\right\}$, as shown in Algorithm 1 in Figure \ref{fig:algorithms}. The rows of $W_{\text{eff}}$ (sparse) give information of which input pixels are influencing a given output pixel. This is of particular interest for classification tasks where transposed networks (also called \emph{deconvolution}) and occlusion sensitivity are a standard visualization techniques \cite{DBLP:conf/eccv/ZeilerF14}. Indeed, we can obtain the rows of $W_{\text{eff}}$ using the transposed network as shown in Algorithm 2 in Figure \ref{fig:algorithms} in the same fashion as \cite{DBLP:conf/eccv/ZeilerF14}. An important difference is that in \cite{DBLP:conf/eccv/ZeilerF14} the activation layers are not fixed and thus their results cannot be interpreted using Theorem \ref{thm:fixed_act}.

\section{Upscaling Systems}
\label{sec:systems}

\subsection{Super--Resolution: Deterministic upscaling}
\label{ssec:sres}
A simple approach to upscale color images is shown in Figure \ref{fig:lq} where the $3$ input color channels are used to estimate the lumminance (only). Color channels are upscaled independetly using standard methods (e.g. bicubic). For the sake of simplicity, the systems in Figure \ref{fig:systems} include noise inputs which will only be useful for HRes methods. In practice, adding noise to the input does not affect the performance of SRes since training methods will adjust weights to stop the noise. This is also useful in practice since we can use identical models for SRes and HRes methods.

\begin{figure}
\noindent\begin{minipage}{\textwidth}
   \centering
   \fbox{\begin{minipage}{.45\textwidth}
     \centering
     \captionof{algorithm}{Forward Analysis} \vspace*{-.03\textwidth}
     \begin{algorithmic}[1]
         \REQUIRE Probe input image $a$.
         \REQUIRE Network parameters $W_n$, $b_n$.
         \REQUIRE Row set $\mathcal{C}\subset\left\{1,\ldots,B\right\}$.

         \STATE // Run network and record activity
         \STATE $x_0 \leftarrow a$
         \FOR{$n=1,\ldots,L$}
             \STATE $y_n \leftarrow W_n x_{n-1} + b_n$
             \STATE $x_n \leftarrow \sigma\left(y_n\right)$
             \STATE $m_n \leftarrow \Sigma(y_n)$
         \ENDFOR

         \STATE // Compute effective bias
         \STATE $z_0 \leftarrow 0$
         \FOR{$n=1,\ldots,L$}
             \STATE $v_n \leftarrow W_n z_{n-1} + b_n$
             \STATE $z_n \leftarrow m_n \cdot v_n$
         \ENDFOR
         \STATE $b_\text{eff} \leftarrow z_L$

         \FOR{$j$ in $\mathcal{C}$}
             \STATE // Run network with
             \STATE // impulse at input $j$
             \STATE $z_0 \leftarrow \delta_j$
             \FOR{$n=1,\ldots,L$}
                 \STATE $v_n \leftarrow W_n z_{n-1}$
                 \STATE $z_n \leftarrow m_n \cdot v_n$
             \ENDFOR
             \STATE $W_\text{eff}[:,\;j] \leftarrow z_L$
         \ENDFOR
         \RETURN $y_L, W_\text{eff}[:,\;\mathcal{C}], b_\text{eff}$
     \end{algorithmic}
   \end{minipage}} \hfil
   \fbox{\begin{minipage}{.45\textwidth}
     \centering
     \captionof{algorithm}{Backward Analysis} \vspace*{-.03\textwidth}
     \begin{algorithmic}[1]
         \REQUIRE Probe input image $a$.
         \REQUIRE Network parameters $W_n$, $b_n$.
         \REQUIRE Column set $\mathcal{R}\subset\left\{1,\ldots,A\right\}$.

         \STATE // Run network and record activity
         \STATE $x_0 \leftarrow a$
         \FOR{$n=1,\ldots,L$}
             \STATE $y_n \leftarrow W_n x_{n-1} + b_n$
             \STATE $x_n \leftarrow \sigma\left(y_n\right)$
             \STATE $m_n \leftarrow \Sigma(y_n)$
         \ENDFOR

         \STATE // Compute effective bias
         \STATE $z_0 \leftarrow 0$
         \FOR{$n=1,\ldots,L$}
             \STATE $v_n \leftarrow W_n z_{n-1} + b_n$
             \STATE $z_n \leftarrow m_n \cdot v_n$
         \ENDFOR
         \STATE $b_\text{eff} \leftarrow z_L$

         \FOR{$i$ in $\mathcal{R}$}
             \STATE // Run transposed network with
             \STATE // impulse at output $i$
             \STATE $z_L \leftarrow \delta_i$
             \FOR{$n=L,\ldots,1$}
                 \STATE $v_{n-1} \leftarrow W^T_n z_n$
                 \STATE $z_{n-1} \leftarrow m_{n-1} \cdot v_{n-1}$
             \ENDFOR
             \STATE $W_\text{eff}[i,\;:] \leftarrow z_0$
         \ENDFOR
         \RETURN $y_L, W_\text{eff}[\mathcal{R},\;:], b_\text{eff}$
     \end{algorithmic}
 \end{minipage}}
\end{minipage}
\caption{Visualization algorithm to obtain parameters $W_\text{eff}$ and $b_\text{eff}$ according to Theorem \ref{thm:fixed_act}.}
\label{fig:algorithms}
\end{figure}

For large upscaling factors there is a risk of misalignment between lumminance and color. This can also be a problem for HRes methods since artificially generated details need consistent alignment between lumminance and color. A simple solution would be to output $3$ features from the network with the $3$ color channels. In practice, this does not work well because color channels are mixed within the network and become hard to untangle for deep networks. Thus, we propose an independent pipeline for the $3$ color channels as shown in Figure \ref{fig:hq}. Finally, in Figure \ref{fig:rq} we proposed our preferred configuration in which we use convolutional networks and Muxout layers to upscale color channels up to one upscaling factor before the full output resolution, and use a simple upscaler (e.g. bicubic) for the last step. This equivalent to the chroma--subsampling approach used in color formats such as YUV--420 in image and video compression \cite{SWinkler_2001a}.

We propose to train these models to maximize the \emph{structural similarity} metric in \cite{Wang04imagequality} which is know to be better correlated to human visual system than the MSE metric, commonly used in SRes methods \cite{CDong_2014a, CDong_2015a}. The loss function is then $L(x) = -\mathbb{E}_{x\sim \mathbb{P}_r}\left[SSIM(x, S_\text{down}(x))\right]$ where $\mathbb{P}_r$ is the distribution of real images in $\mathcal{H}^R$.

\subsection{Hyper--Resolution: Generative upscalers}
\label{ssec:hres}
When upscaling factors are large, the SRes approach of finding \emph{one} particular image in $\mathcal{A}^R$, that is never in the training samples $\mathcal{S}$, seems increasingly difficult. Moreover, optimizing image quality metrics makes it impossible for SRes methods to reach an element of $\mathcal{A}^R$ because of the highly random nature of the missing details. These detais correspond to an \emph{innovation process} that is cancelled by expected values in the loss functions. For applications such as digital television, HRes offers a more feasible approach by searching for \emph{any} image in $\mathcal{A}^R$. This is an acceptable solution when $\mathcal{L}$ corresponds to standard--definition resolutions because the missing details will typically correspond to textures (e.g. skin, grass, hair, etc.) that do not need to be identical to the original.

\begin{figure}
    \centering
    \subfloat[Low color quality configuration.]{\includegraphics[width=0.493\linewidth]{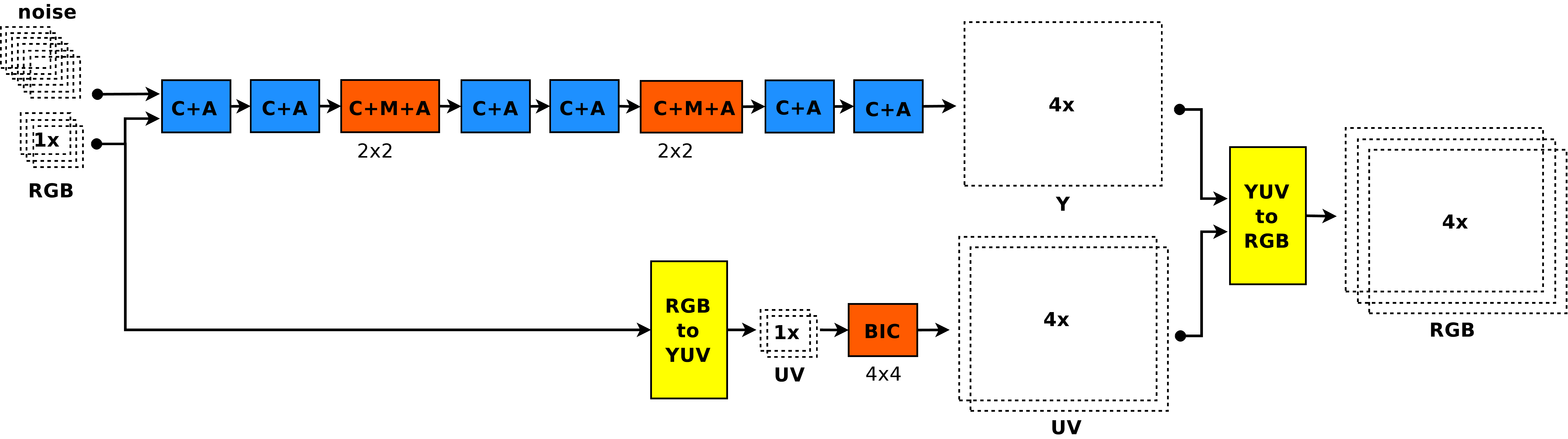} \label{fig:lq}} \hfil
    \subfloat[High color quality configuration.]{\includegraphics[width=0.493\linewidth]{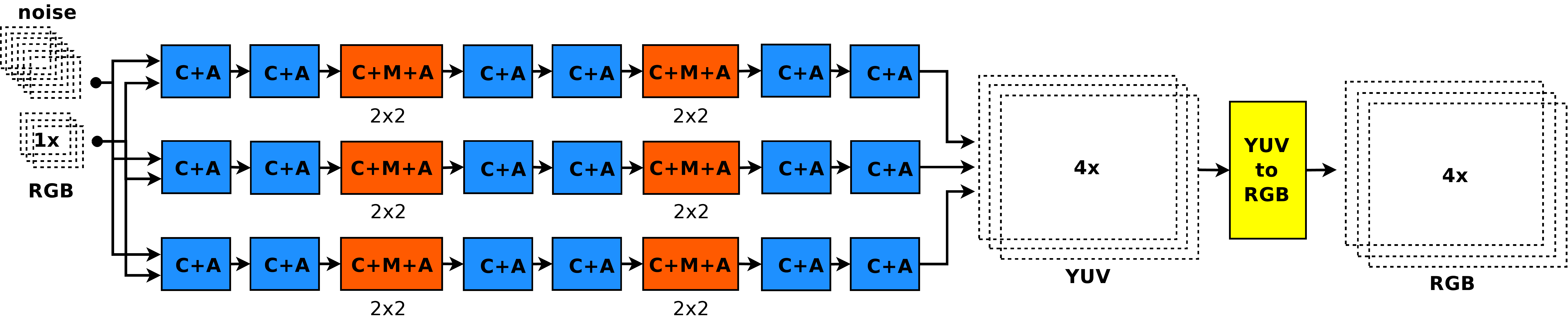} \label{fig:hq}} \hfil
    \centering
    \subfloat[Chroma sub--sampling (preferred) configuration.]{\includegraphics[width=0.493\linewidth]{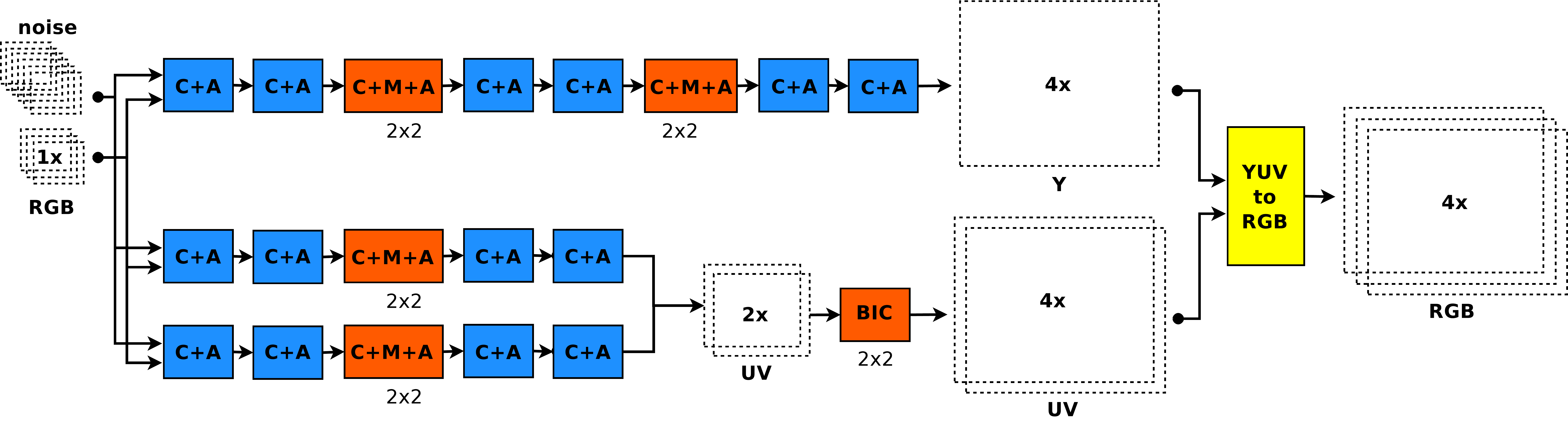} \label{fig:rq}} \hfil
    \subfloat[Discriminator used for GAN training.]{\includegraphics[width=0.493\linewidth]{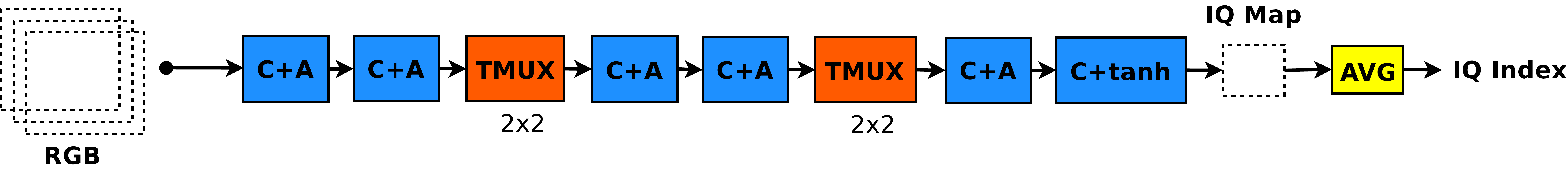} \label{fig:disc}}
    \caption{Systems architectures for SRes and HRes upscalers.}
    \label{fig:systems}
\end{figure}

\emph{Generative Adversarial Networks} (GANs) allow us to use the same system architecture of SRes methods in Figure \ref{fig:systems} to configure HRes methods. Now, the noise inputs are necessary in order to approximate a distribution of artificially generated details that for a fixed noise input gives us particular fixed details. The noise is manipulated to generate the \emph{innovation process} that allows us to get an actual image from $\mathcal{A}^R$. For the training process we use the \emph{Discriminator} system, $D$, shown in Figure \ref{fig:disc}. A final \emph{tanh} activation layer forces an output value between $-1$, meaning that the input image is the output of the upscaler (fake), and $1$, meaning that the input image is original unmodified content (real). Here, we choose to configure a system that is symmetric with respect to the upscaler. Thus, although not strictly necessary, we use T--MuxOut as pooling layer. The training process will then alternate steps to: improve the upscaler to increase the outputs of $D$, and improve the discriminator to better identify which images are upscaled and which are not.

Training GANs is an extensive topic of research and here we choose the particular approach of \emph{Wasserstein GAN} (WGAN) \cite{2017arXiv170107875A} with the improvement reported in \cite{2017arXiv170400028G} to train our system. This is, we train the Discriminator and our upscaler (Generator) using the following loss functions:
\begin{eqnarray}
    L_D & = & \mathbb{E}_{\tilde{x}\sim \mathbb{P}_g}\left[D(\tilde{x})\right] - \mathbb{E}_{x\sim \mathbb{P}_r}\left[D(x)\right] + \lambda_{gp}\; \mathbb{E}_{\hat{x}\sim \mathbb{P}_{\hat{x}}}\left[\left(\lVert \nabla_{\hat{x}}D(\hat{x})\rVert_2-1 \right)^2\right]\\
    L_G & = & -\mathbb{E}_{\tilde{x}\sim \mathbb{P}_g}\left[D(\tilde{x})\right] + \lambda_\text{down} \; \mathbb{E}_{\breve{x}\sim \mathbb{P}_\text{LR}}\left[\Delta\left(\breve{x}, S_\text{down}(G(\breve{x}))\right)\right]
\end{eqnarray}
where $\mathbb{P}_g$ and $\mathbb{P}_r$ are the distributions of upscaled and real high--resolution images, respectively; $\hat{x}=\epsilon x + (1-\epsilon)\tilde{x}$ with $\epsilon \sim U[0,1],\; x\sim \mathbb{P}_r \;, \tilde{x}\sim \mathbb{P}_g$; and $\mathbb{P}_\text{LR}$ is the distributions of realistic low--resolution images in $\mathcal{L}^R$. The metric $\Delta$ can be chosen as $\Delta\left(x,y\right) = MSE(x,y)$ or $\Delta\left(x,y\right) = 1-SSIM(x,y)$. Here, we intentionally avoid high--resolution content in the generator (upscaler) loss to force the upscaler to look for \emph{any} and not just \emph{one} image in $\mathcal{A}^R$. Specifically, we force the upscaler to be able to recover the low--resolution input with a standard downscaler (e.g. area) since otherwise it might just ignore the low--resolution input and output constant images (the simplest realistic images). This approach follows the spirit of the so--called Cycle--GAN \cite{CycleGAN2017} and has been reported both in \cite{FaceSR} and \cite{PNavarrete_2017a}.

\begin{figure}[h]
    \centerline{
    \subfloat[Standard (PSNR 24.82 dB -- SSIM 0.8463)]{\includegraphics[width=0.5\textwidth]{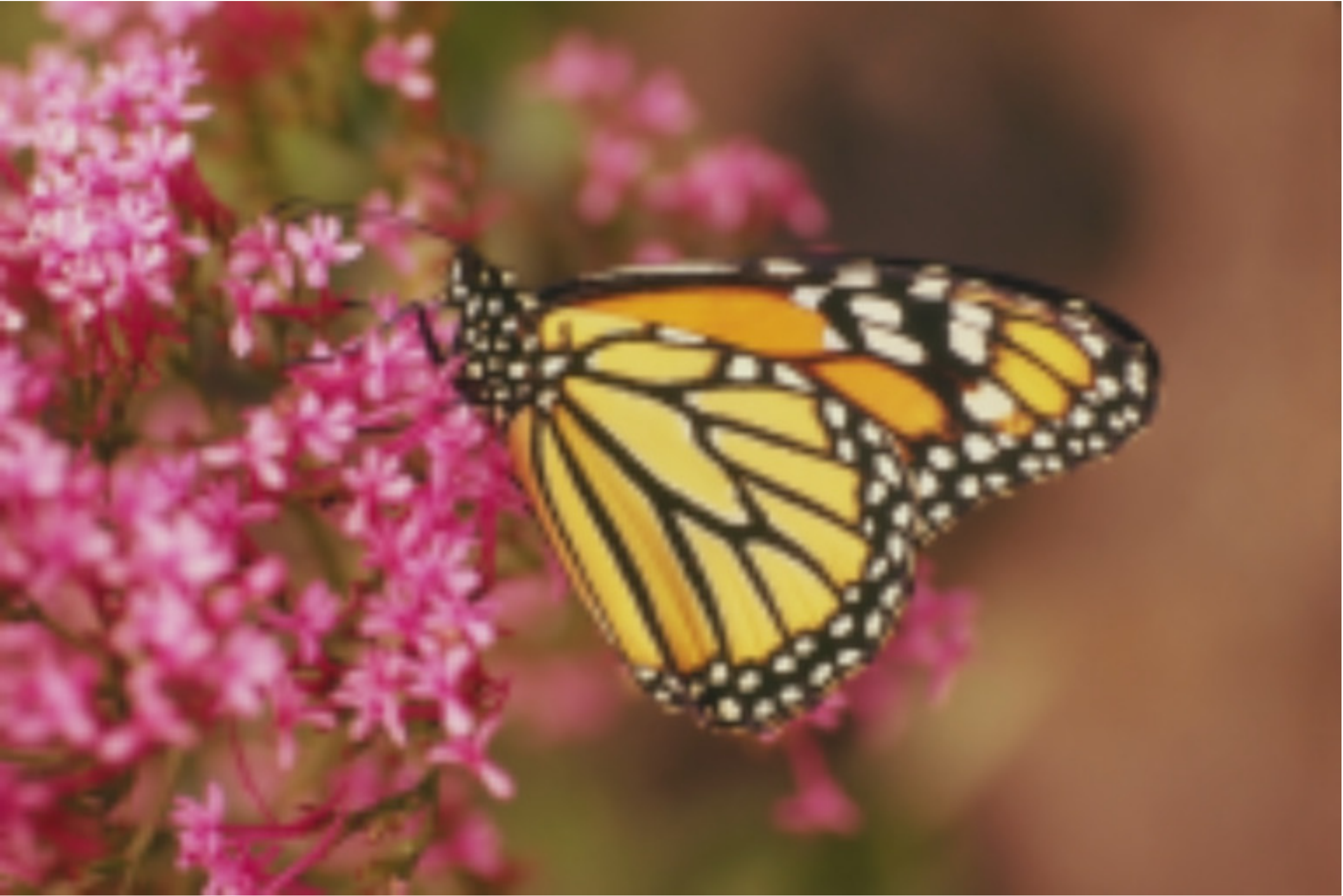} \label{fig:ex1_std}} \hfil
    \subfloat[SRes output (PSNR 27.31 dB -- SSIM 0.8990)]{\includegraphics[width=0.5\textwidth]{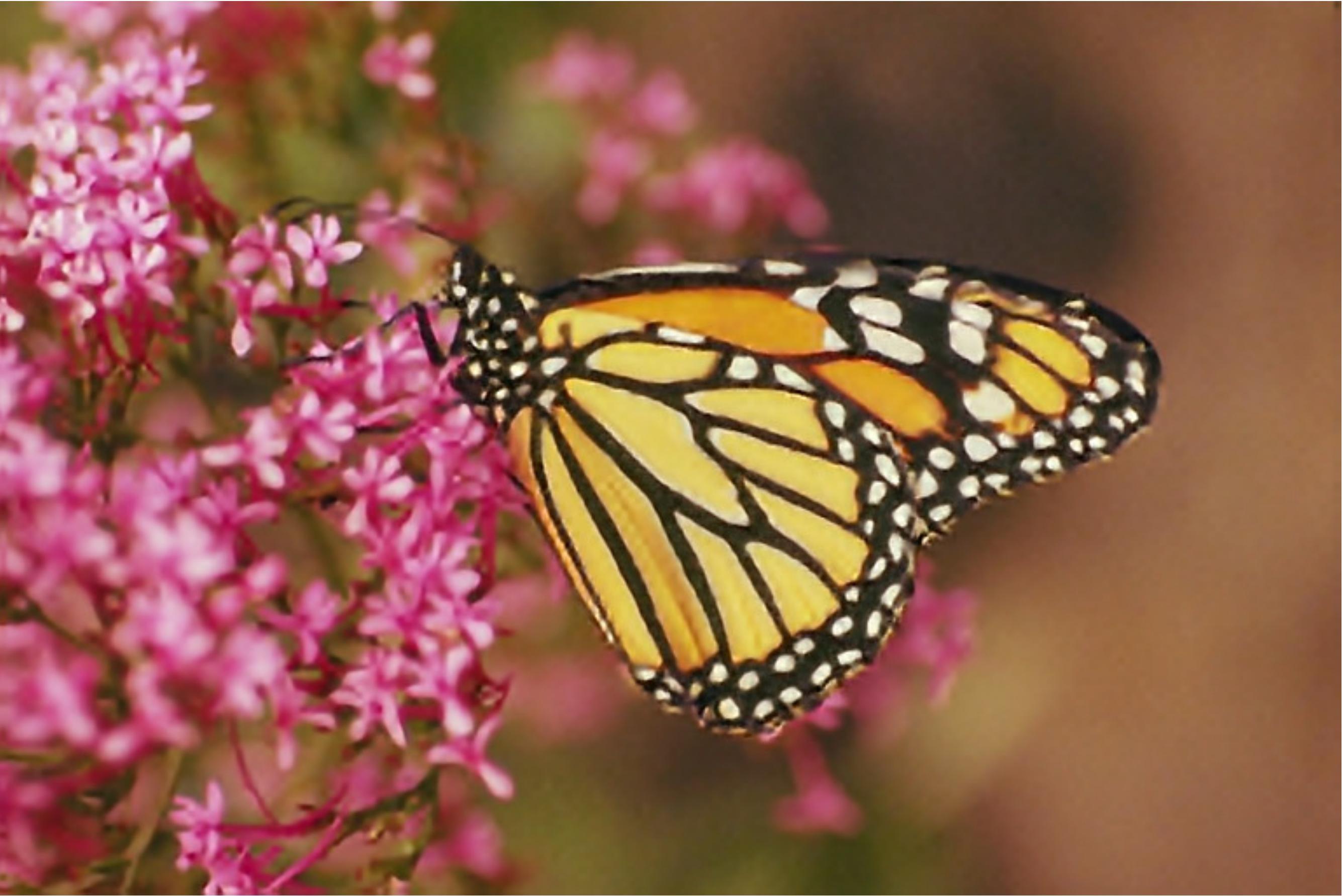} \label{fig:ex1_ssim}}
    }
\end{figure}

\begin{figure}[h]
    \hspace*{-0.02\textwidth}
    \subfloat[Standard (PSNR 29.78 dB)]{ \label{fig:ex2_std}
    \begin{tabular}{c}
        \includegraphics[width=0.31\textwidth]{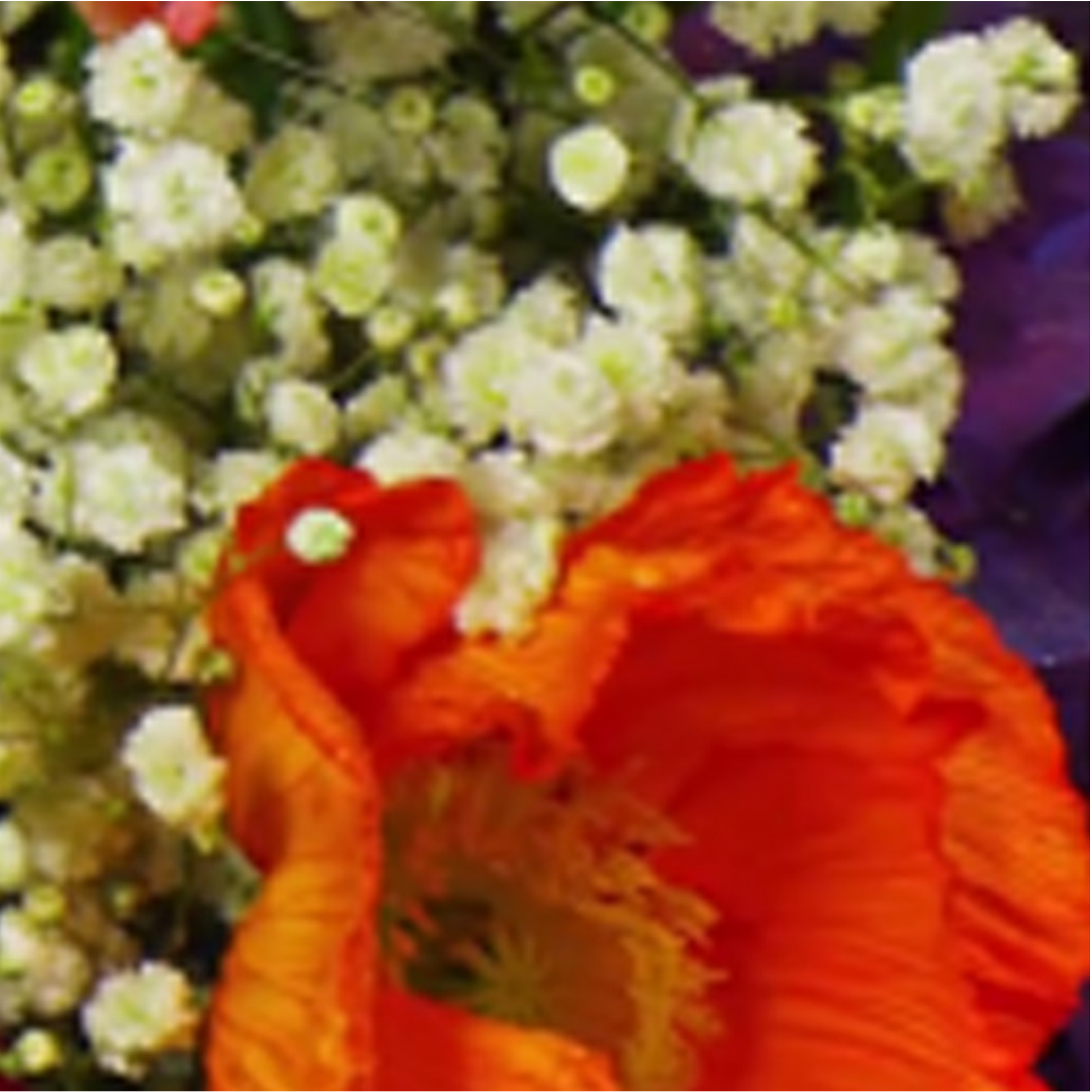} \\
        \includegraphics[width=0.31\textwidth]{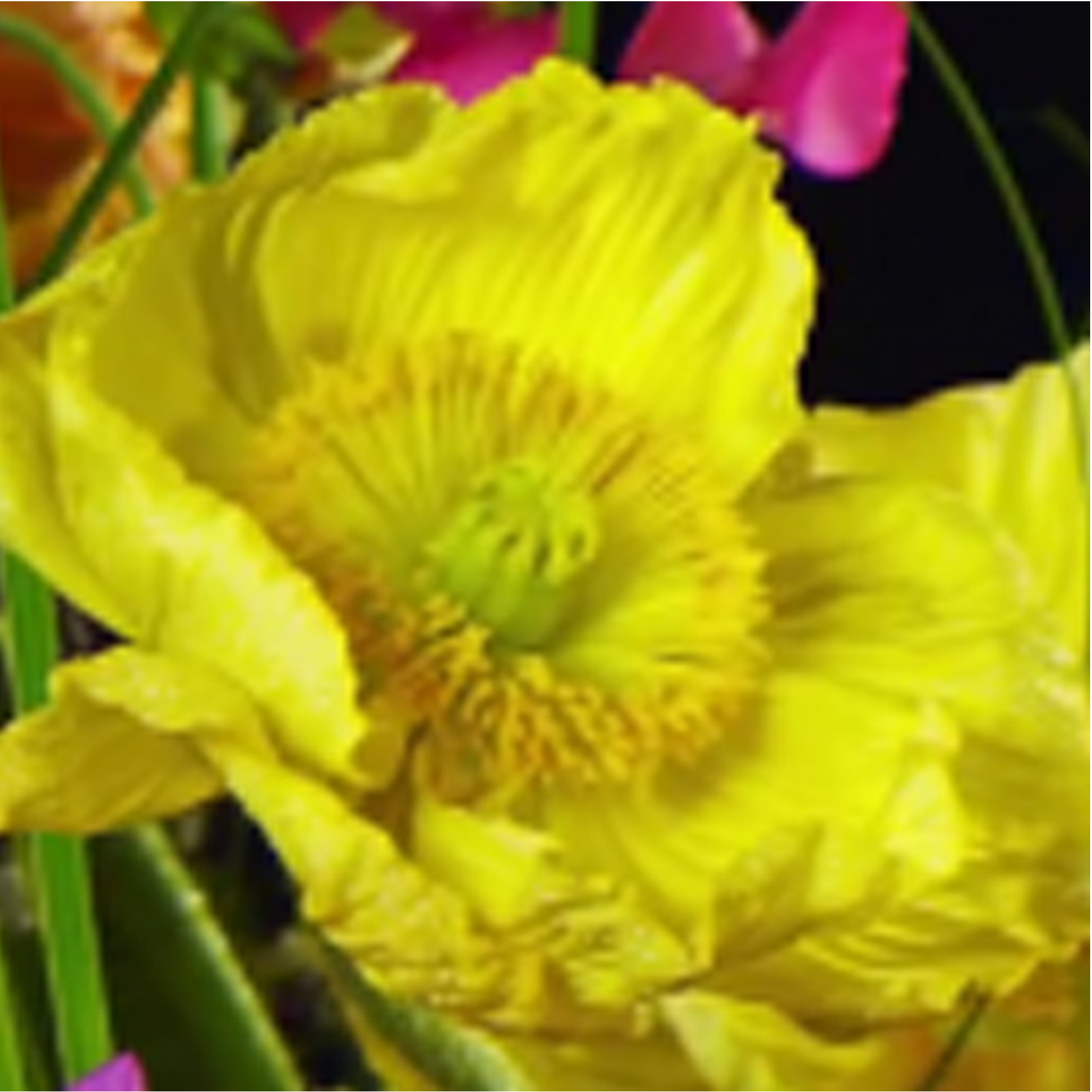}
    \end{tabular}
    }
    \subfloat[Original (PSNR $\infty$)]{ \label{fig:ex2_original}
    \begin{tabular}{c}
        \includegraphics[width=0.31\textwidth]{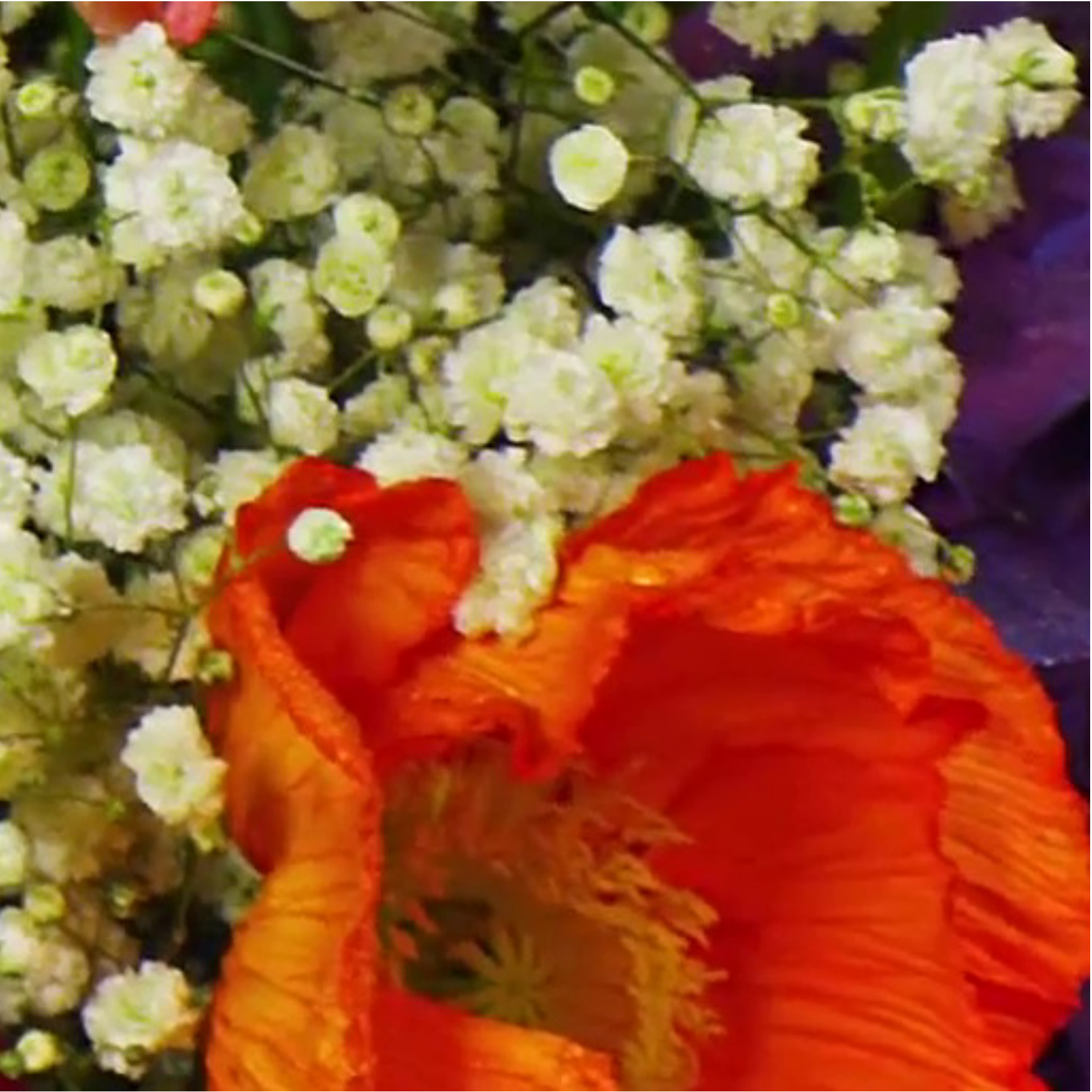} \\
        \includegraphics[width=0.31\textwidth]{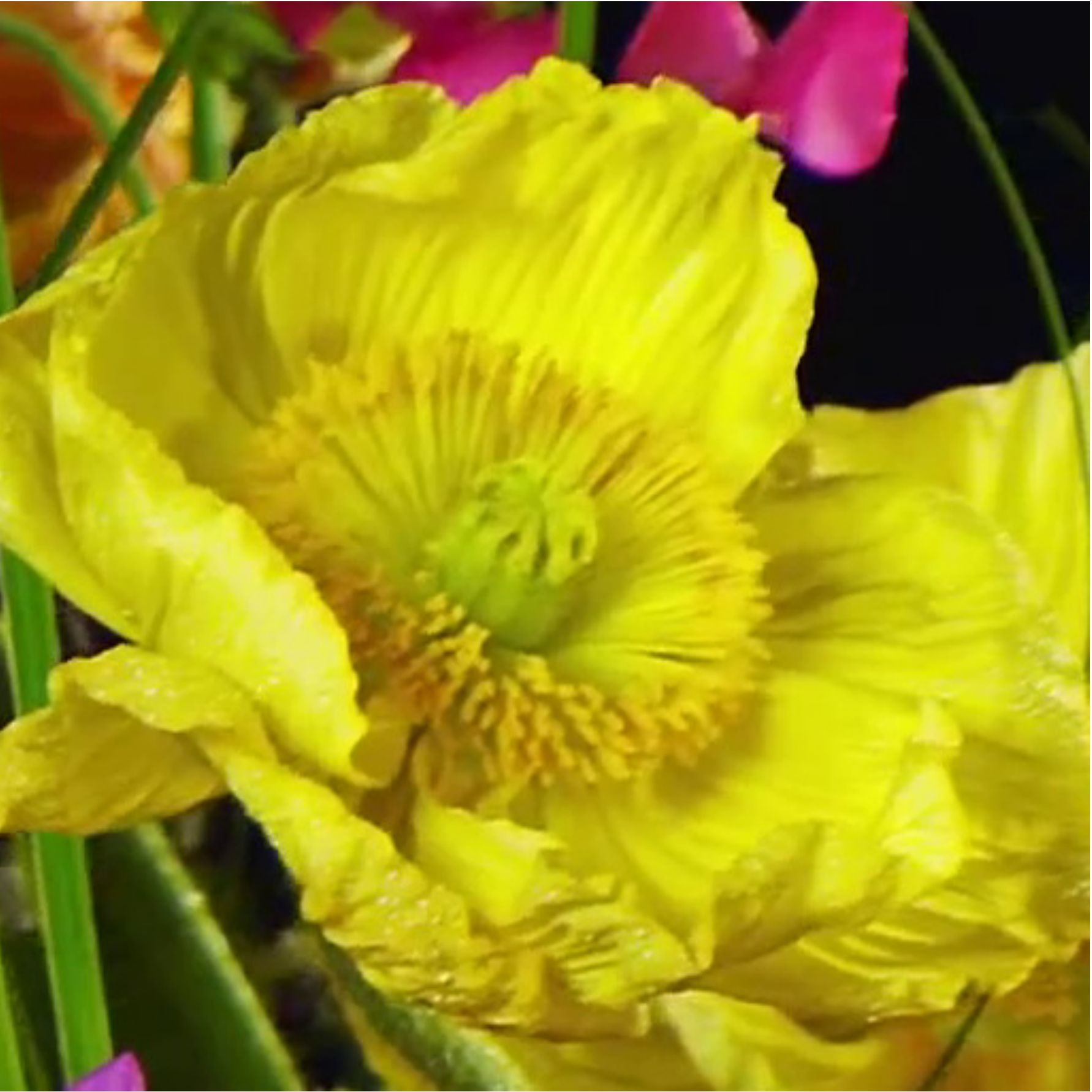}
    \end{tabular}
    }
    \subfloat[HRes output (PSNR 25.68 dB)]{ \label{fig:ex2_wgan}
    \begin{tabular}{c}
        \includegraphics[width=0.31\textwidth]{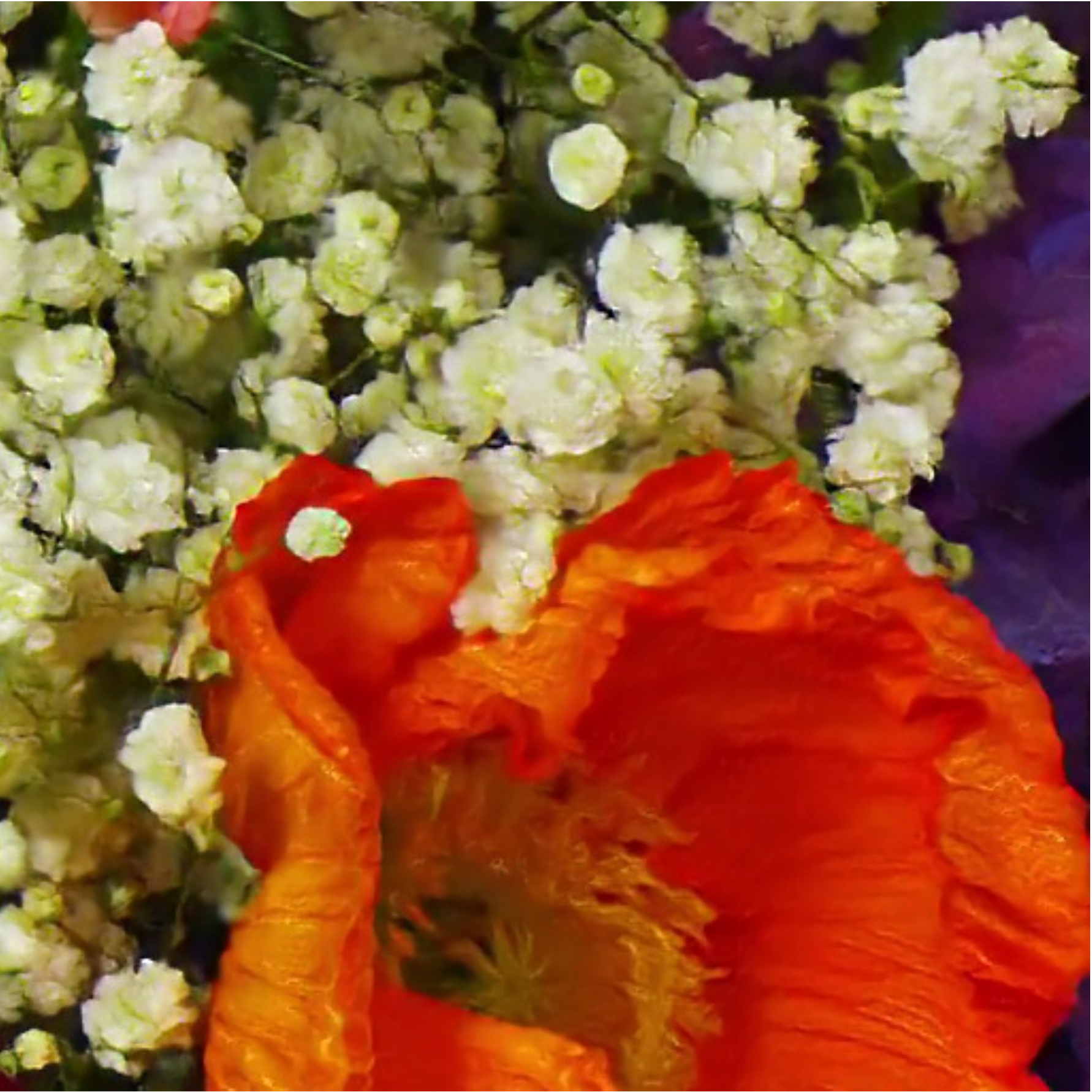} \\
        \includegraphics[width=0.31\textwidth]{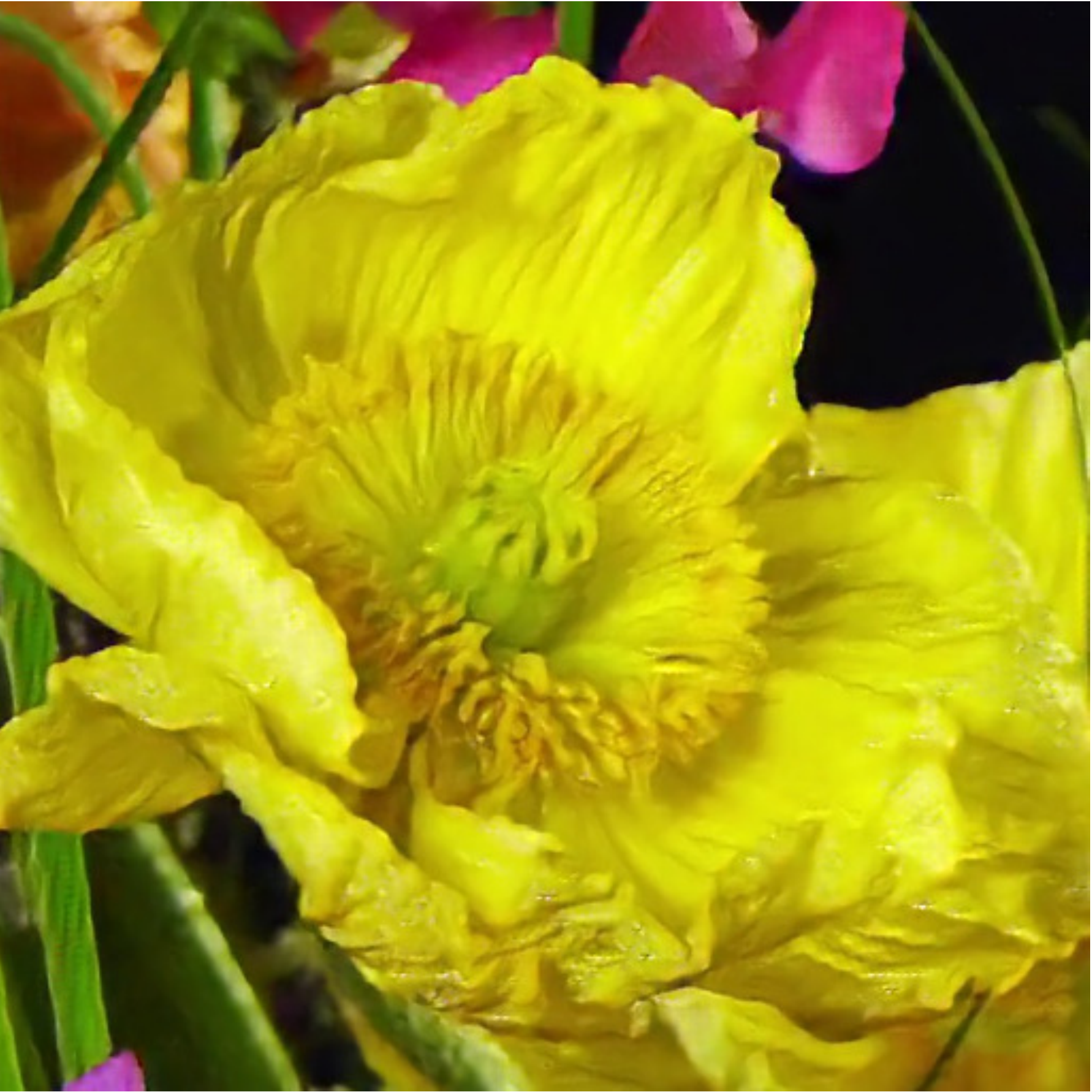}
    \end{tabular}
}
\end{figure}

\begin{figure}[h]
    \label{fig:analysis}
    \subfloat[Input]{\includegraphics[width=0.2\textwidth]{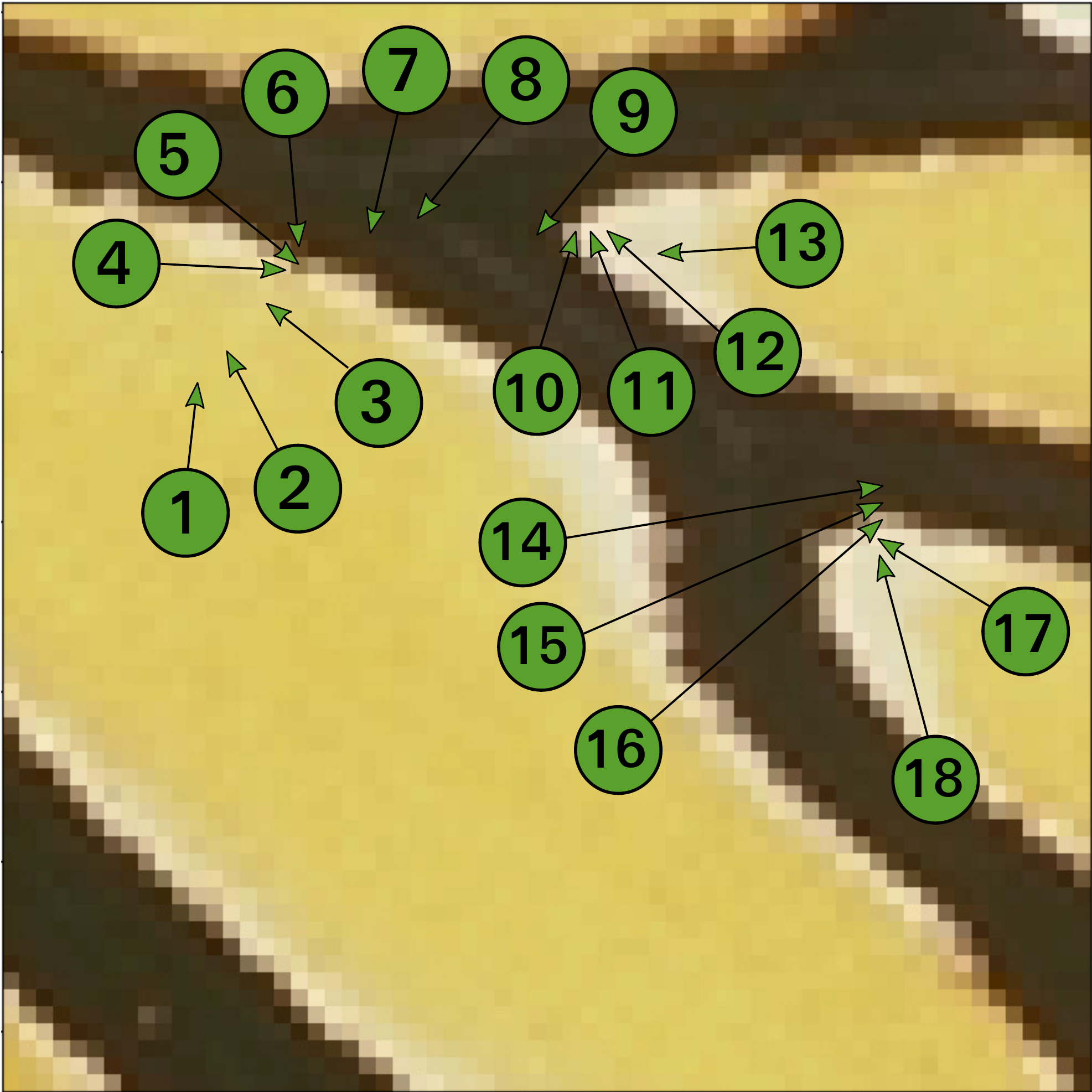} \label{fig:analysis_input}} \hfil
    \subfloat[SRes output $Y$]{\includegraphics[width=0.19\textwidth]{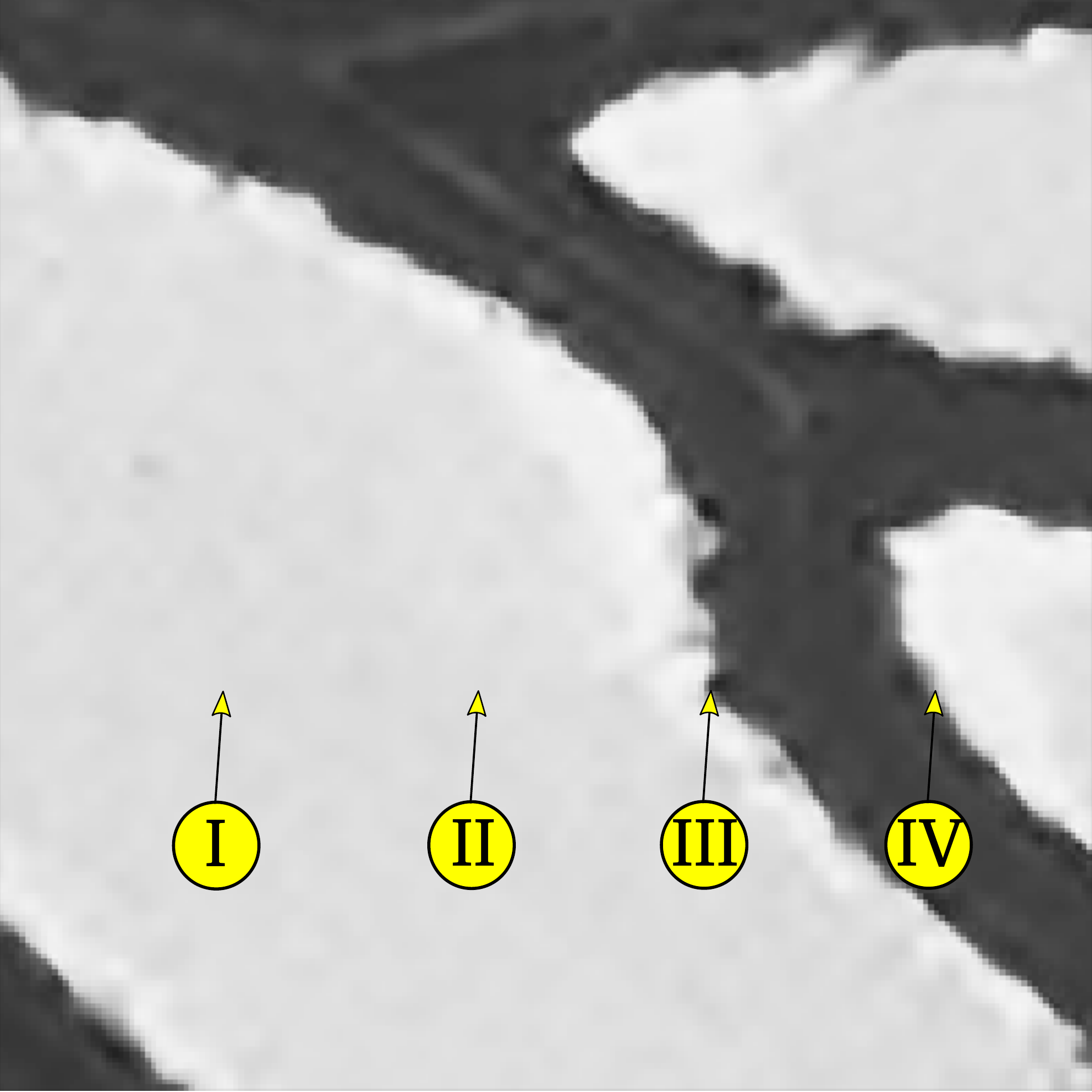} \label{fig:analysis_sres_out}} \hfil
    \subfloat[HRes output $Y$]{\includegraphics[width=0.19\textwidth]{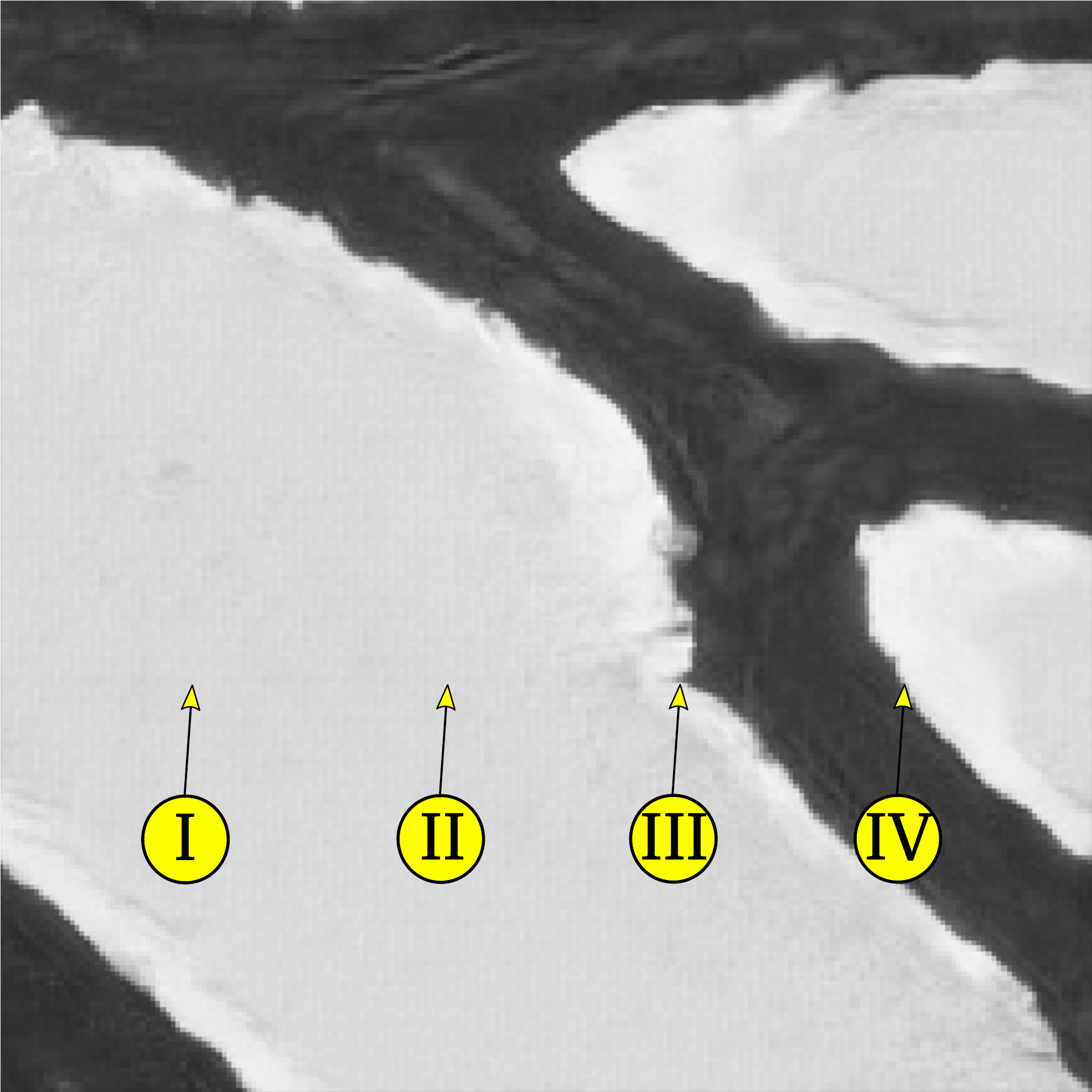} \label{fig:analysis_hres_out}} \\
    \centering
    \subfloat[Columns of $W_\text{eff}$ (adaptive interpolation coefficients) for SRes at input locations in Figure \ref{fig:analysis_input}]{\fbox{\begin{tabular}{c@{\hskip 0.005\textwidth}c@{\hskip 0.005\textwidth}c@{\hskip 0.005\textwidth}c@{\hskip 0.005\textwidth}c@{\hskip 0.005\textwidth}c@{\hskip 0.005\textwidth}c@{\hskip 0.005\textwidth}c@{\hskip 0.005\textwidth}c}
        \vspace*{-0.01\textwidth}
        \includegraphics[width=0.1\textwidth]{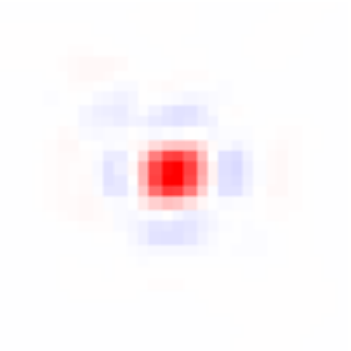} &
        \includegraphics[width=0.1\textwidth]{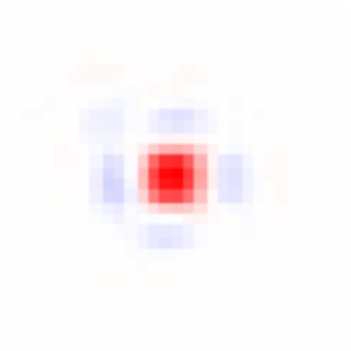} &
        \includegraphics[width=0.1\textwidth]{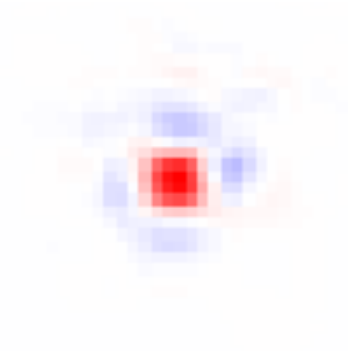} &
        \includegraphics[width=0.1\textwidth]{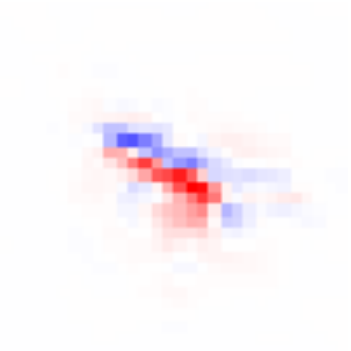} &
        \includegraphics[width=0.1\textwidth]{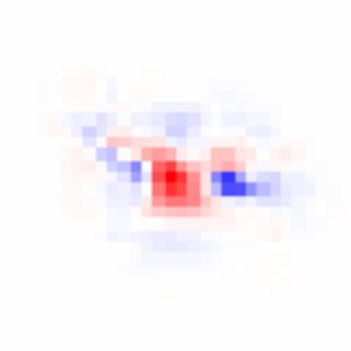} &
        \includegraphics[width=0.1\textwidth]{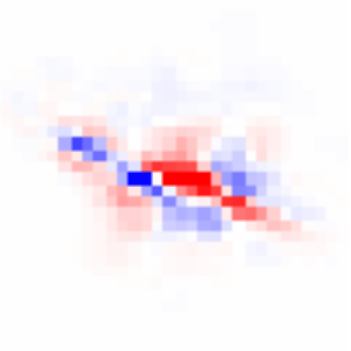} &
        \includegraphics[width=0.1\textwidth]{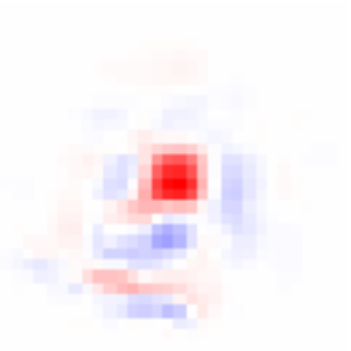} &
        \includegraphics[width=0.1\textwidth]{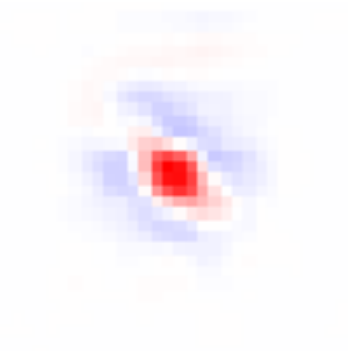} &
        \includegraphics[width=0.1\textwidth]{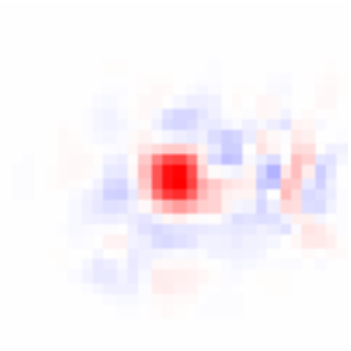} \\
        \circled{1} &
        \circled{2} &
        \circled{3} &
        \circled{4} &
        \circled{5} &
        \circled{6} &
        \circled{7} &
        \circled{8} &
        \circled{9} \\
        \vspace*{-0.01\textwidth}
        \includegraphics[width=0.1\textwidth]{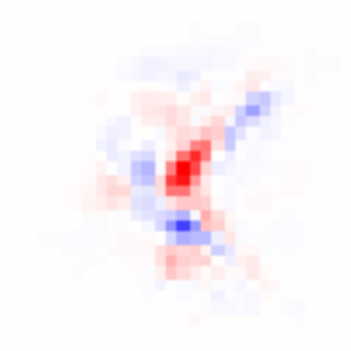} &
        \includegraphics[width=0.1\textwidth]{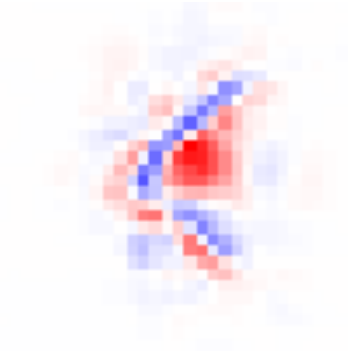} &
        \includegraphics[width=0.1\textwidth]{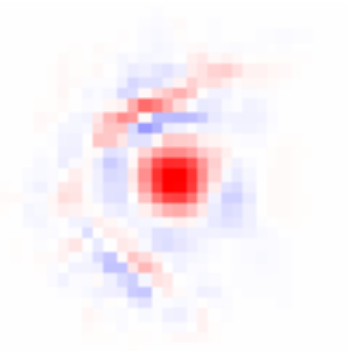} &
        \includegraphics[width=0.1\textwidth]{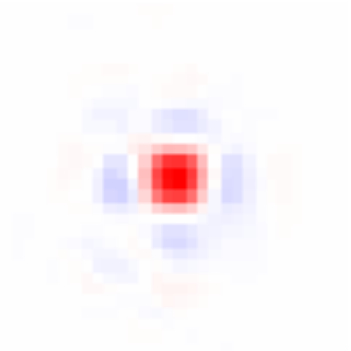} &
        \includegraphics[width=0.1\textwidth]{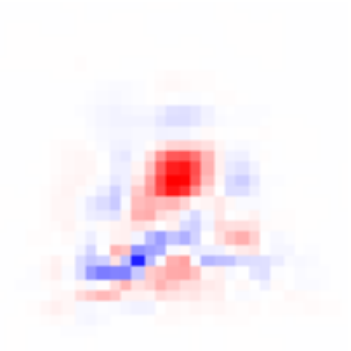} &
        \includegraphics[width=0.1\textwidth]{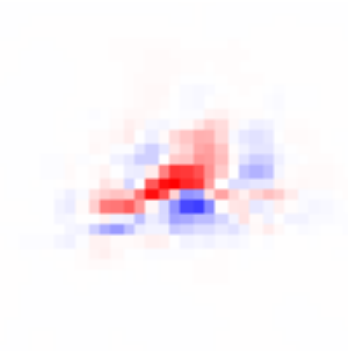} &
        \includegraphics[width=0.1\textwidth]{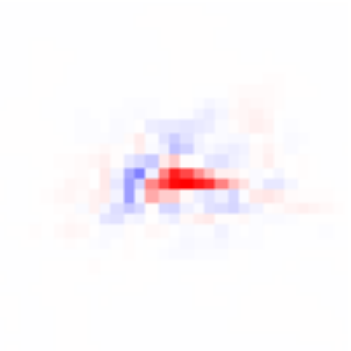} &
        \includegraphics[width=0.1\textwidth]{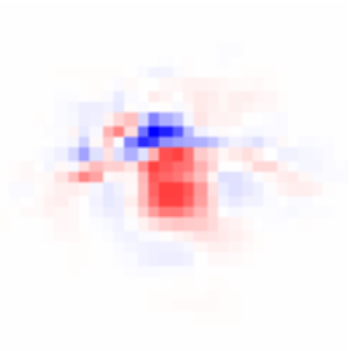} &
        \includegraphics[width=0.1\textwidth]{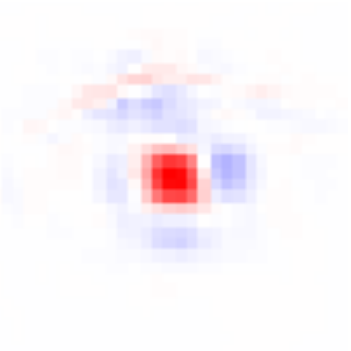} \\
        \circled{10} &
        \circled{11} &
        \circled{12} &
        \circled{13} &
        \circled{14} &
        \circled{15} &
        \circled{16} &
        \circled{17} &
        \circled{18}
    \end{tabular}} \label{fig:analysis_cols_sres}} \\
    \centering
    \subfloat[Columns of $W_\text{eff}$ (adaptive interpolation coefficients) for HRes at input locations in Figure \ref{fig:analysis_input}]{\fbox{\begin{tabular}{c@{\hskip 0.005\textwidth}c@{\hskip 0.005\textwidth}c@{\hskip 0.005\textwidth}c@{\hskip 0.005\textwidth}c@{\hskip 0.005\textwidth}c@{\hskip 0.005\textwidth}c@{\hskip 0.005\textwidth}c@{\hskip 0.005\textwidth}c}
        \vspace*{-0.01\textwidth}
        \includegraphics[width=0.1\textwidth]{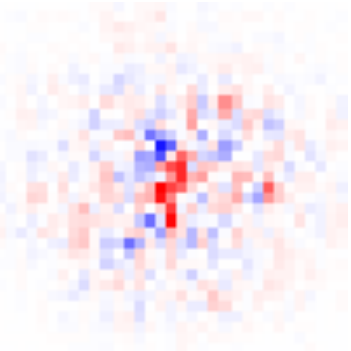} &
        \includegraphics[width=0.1\textwidth]{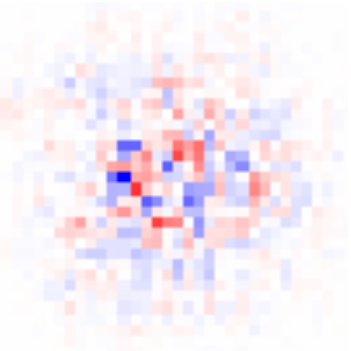} &
        \includegraphics[width=0.1\textwidth]{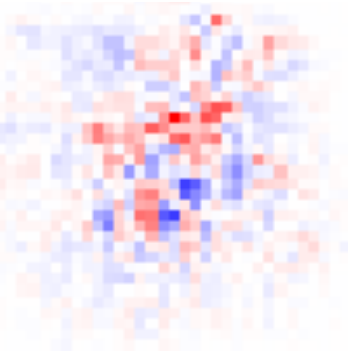} &
        \includegraphics[width=0.1\textwidth]{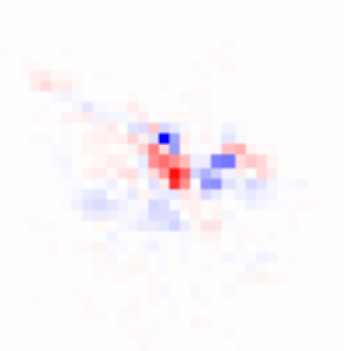} &
        \includegraphics[width=0.1\textwidth]{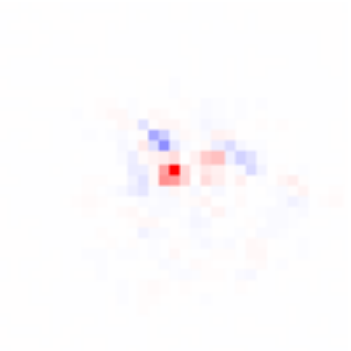} &
        \includegraphics[width=0.1\textwidth]{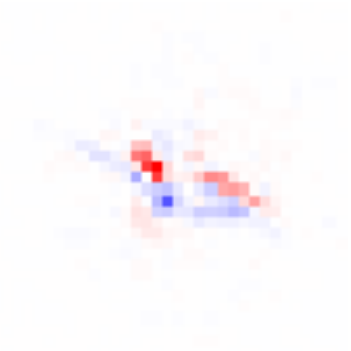} &
        \includegraphics[width=0.1\textwidth]{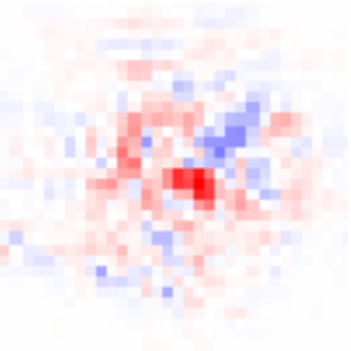} &
        \includegraphics[width=0.1\textwidth]{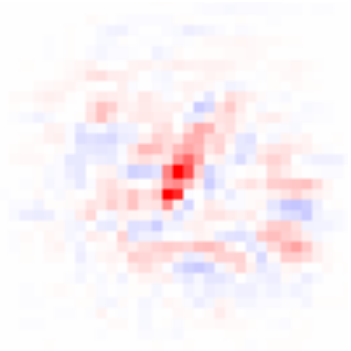} &
        \includegraphics[width=0.1\textwidth]{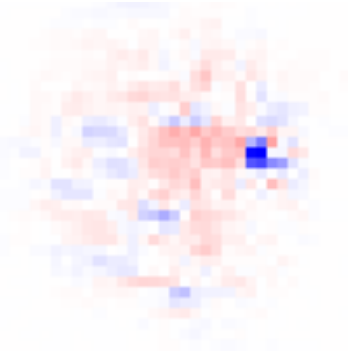} \\
        \circled{1} &
        \circled{2} &
        \circled{3} &
        \circled{4} &
        \circled{5} &
        \circled{6} &
        \circled{7} &
        \circled{8} &
        \circled{9} \\
        \vspace*{-0.01\textwidth}
        \includegraphics[width=0.1\textwidth]{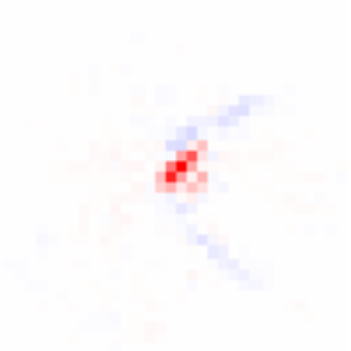} &
        \includegraphics[width=0.1\textwidth]{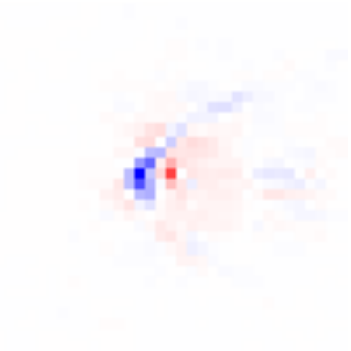} &
        \includegraphics[width=0.1\textwidth]{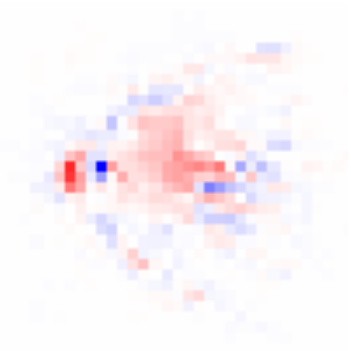} &
        \includegraphics[width=0.1\textwidth]{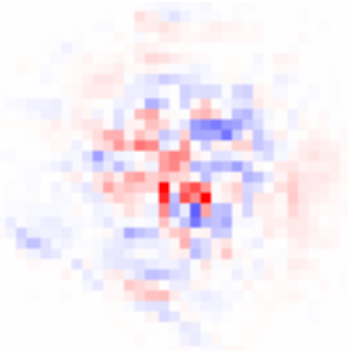} &
        \includegraphics[width=0.1\textwidth]{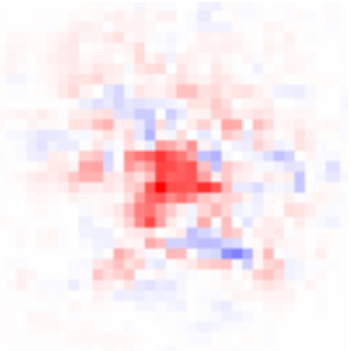} &
        \includegraphics[width=0.1\textwidth]{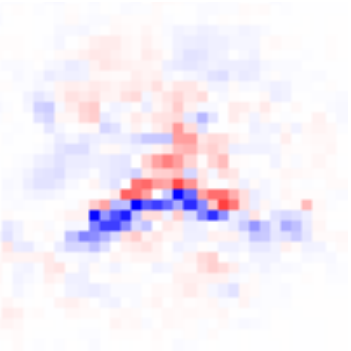} &
        \includegraphics[width=0.1\textwidth]{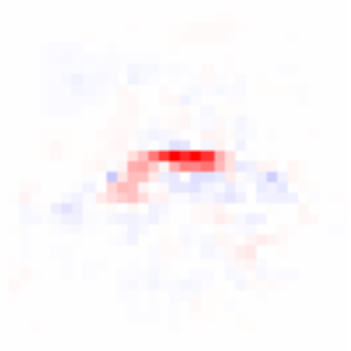} &
        \includegraphics[width=0.1\textwidth]{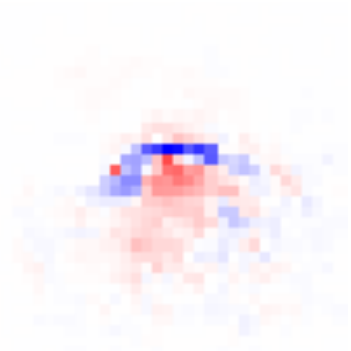} &
        \includegraphics[width=0.1\textwidth]{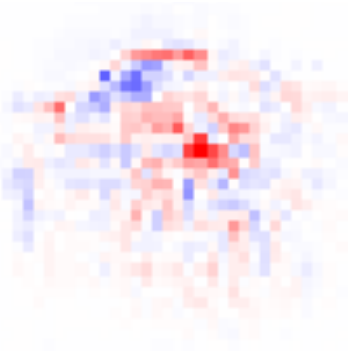} \\
        \circled{10} &
        \circled{11} &
        \circled{12} &
        \circled{13} &
        \circled{14} &
        \circled{15} &
        \circled{16} &
        \circled{17} &
        \circled{18}
    \end{tabular}} \label{fig:analysis_cols_hres}} \\
    \centering
    \subfloat[$b_\text{eff}$ for SRes]{\includegraphics[width=0.19\textwidth]{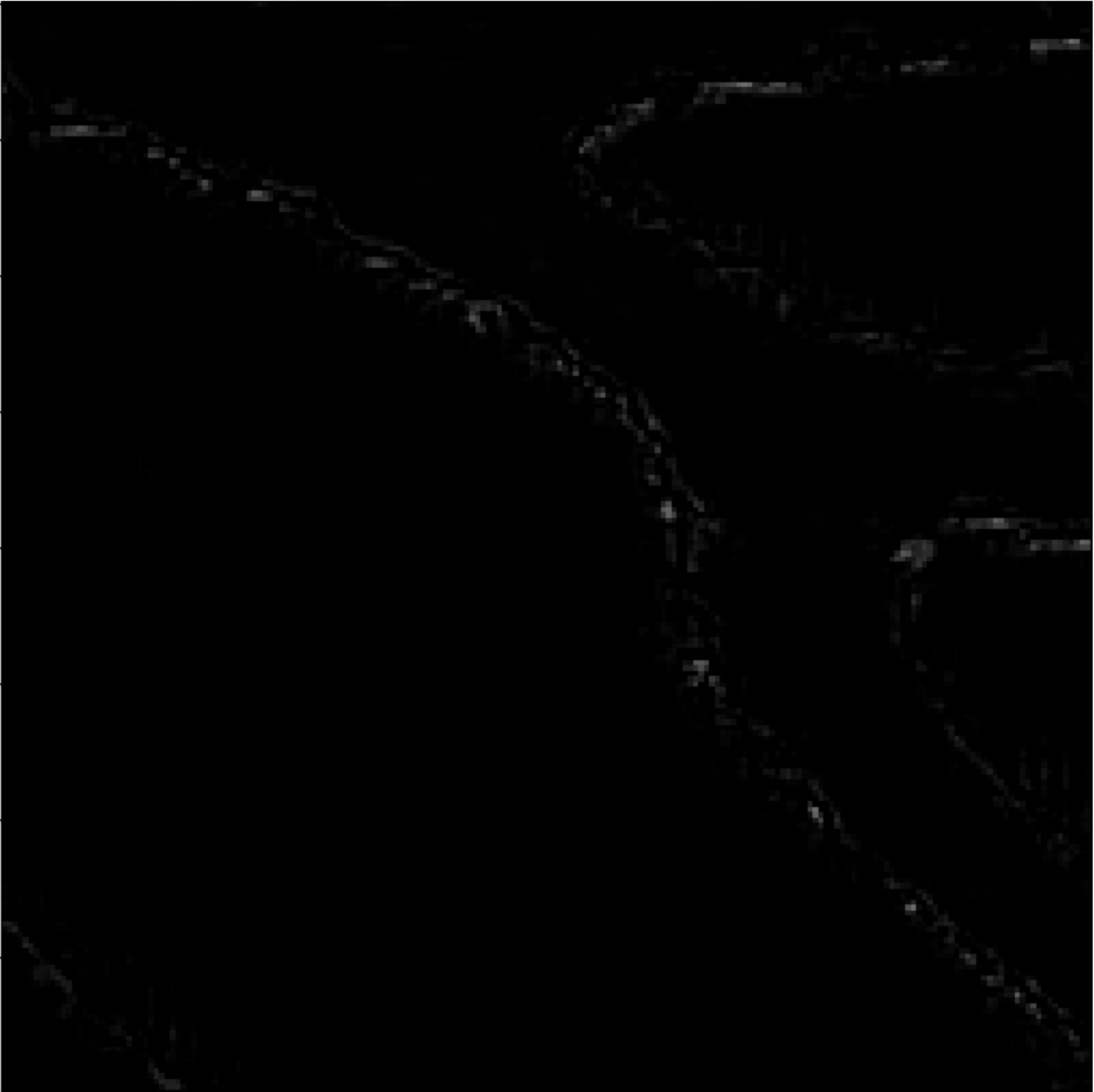} \label{fig:analysis_beff_sres}} \hfil
    \raisebox{0.07\linewidth}{\subfloat[Rows of $W_\text{eff}$ for SRes at output locations in Figure \ref{fig:analysis_sres_out}]{\fbox{\begin{tabular}{c@{\hskip 0.005\textwidth}c@{\hskip 0.005\textwidth}c@{\hskip 0.005\textwidth}c@{\hskip 0.005\textwidth}c@{\hskip 0.005\textwidth}c@{\hskip 0.005\textwidth}c@{\hskip 0.005\textwidth}c@{\hskip 0.005\textwidth}c}
        \vspace*{-0.01\textwidth}
        \includegraphics[width=0.05\textwidth]{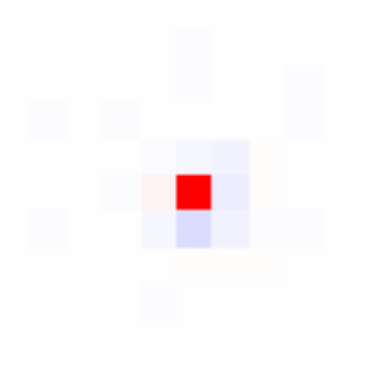} &
        \includegraphics[width=0.05\textwidth]{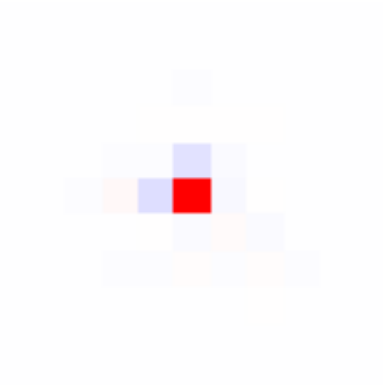} &
        \includegraphics[width=0.05\textwidth]{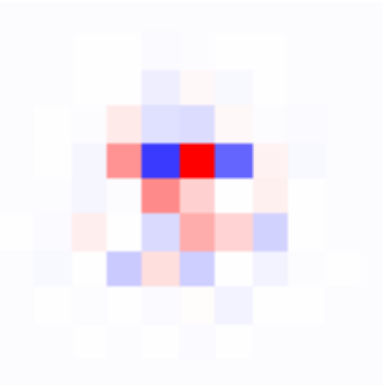} &
        \includegraphics[width=0.05\textwidth]{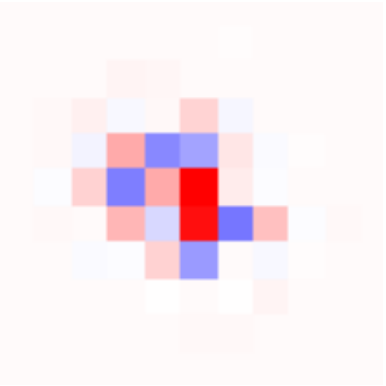} & \\
        \circled{I} &
        \circled{II} &
        \circled{III} &
        \circled{IV}
    \end{tabular}} \label{fig:analysis_rows_sres}}} \hfil
    \subfloat[$b_\text{eff}$ for HRes]{\includegraphics[width=0.19\textwidth]{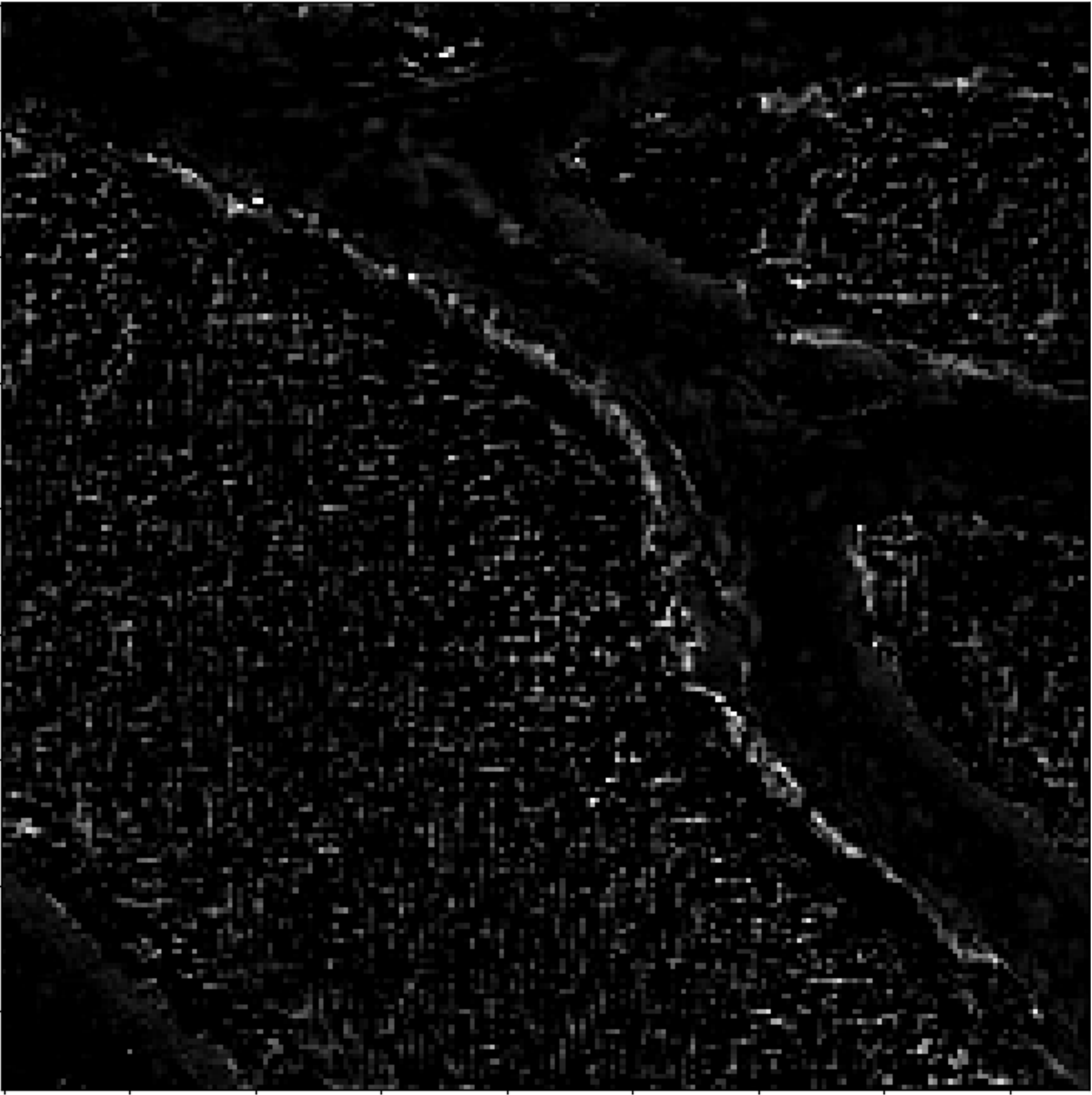} \label{fig:analysis_beff_hres}} \hfil
    \raisebox{0.07\linewidth}{\subfloat[Rows of $W_\text{eff}$ for HRes at output locations in Figure \ref{fig:analysis_hres_out}]{\fbox{\begin{tabular}{c@{\hskip 0.005\textwidth}c@{\hskip 0.005\textwidth}c@{\hskip 0.005\textwidth}c@{\hskip 0.005\textwidth}c@{\hskip 0.005\textwidth}c@{\hskip 0.005\textwidth}c@{\hskip 0.005\textwidth}c@{\hskip 0.005\textwidth}c}
        \vspace*{-0.01\textwidth}
        \includegraphics[width=0.05\textwidth]{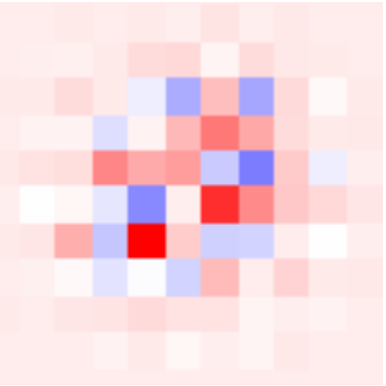} &
        \includegraphics[width=0.05\textwidth]{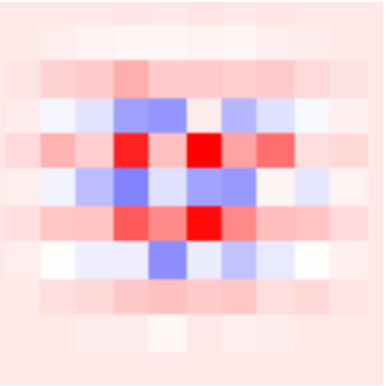} &
        \includegraphics[width=0.05\textwidth]{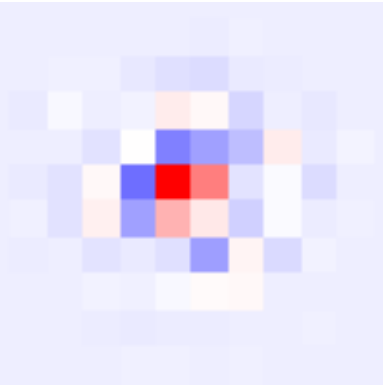} &
        \includegraphics[width=0.05\textwidth]{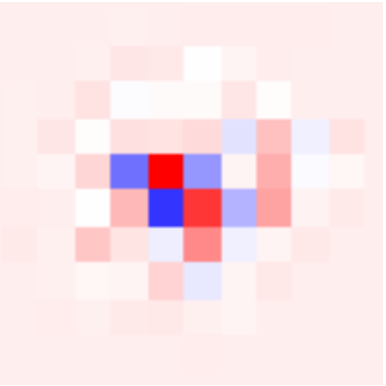} \\
        \circled{I} &
        \circled{II} &
        \circled{III} &
        \circled{IV}
    \end{tabular}} \label{fig:analysis_rows_hres}}}
    \caption{Effective filters and bias obtained with our visualization algorithms.}
\end{figure}

\section{Experiments}
\label{sec:experiments}

We used the system architecture in Figure \ref{fig:rq} for SRes and HRes configurations. We fix the numbers of features to $64$ for all layers except input and output. We train SRes to maximize an SSIM metric as specified in section \ref{ssec:sres} using Adam optimizer with learning rate $10^{-3}$, a training set of $30,000$ patches randomly selected from BSDS500 dataset \cite{amfm_pami2011}, and mini--batch size $20$. For HRes we use Adam optimizer with learning rate $10^{-4}$, minibatch size $20$ for discriminator and generator, we regenerate samples every iteration and set all other hyper--parameters as specified in \cite{2017arXiv170400028G}.

In Figures \ref{fig:ex1_std} and \ref{fig:ex1_ssim} we show an example of SRes output compared to a standard upscaler. We observe sharp edges and smooth curves with an improvement in both PSNR and SSIM metrics. In Figures \ref{fig:ex2_std}, \ref{fig:ex2_original} and \ref{fig:ex2_wgan} we show snapshots of 4K images obtained from a standard $4\times 4$ upscaler, the original (unmodified) image, and our HRes upscaler, respectively. We observe that despite the lower PSNR in our output, our HRes $4\times 4$ upscaled images look as real as the original. The lower PSNR is due to different details that are visible between Figures \ref{fig:ex2_original} and \ref{fig:ex2_wgan}. This proves that our HRes upscaler has been able to obtain samples of $\mathcal{A}^R$ that are different than the original content.

Finally, it is hard to prove the learning ability of the network. Super--resolution research has been focused on the analysis of image quality metrics but with HRes approaches we see improved subjective quality and lower objective quality metrics. We can then observe the adaptive interpolation coefficients at different input locations in Figures \ref{fig:analysis_cols_sres} and \ref{fig:analysis_cols_hres}. For SRes we observe isotropic filters in flat areas, similar to classic upscalers, and directional filters as we approach edges. The effective filter is highly adaptive to the geometry even in complicated curves. For HRes we also observe directional filters as we approach edges, but in flat areas the filters are highly random. This is because in flat areas our HRes system is trying to generate textures which are completely artificial. Thus, we are able to visualize the conservative approach of SRes to avoid an innovation process that might increase the loss, compared to the creative approach of HRes to generate the innovation process but at the same time stay close to the geometry of the input image.

\section{Conclusions}
We proposed a convolutional network structure for image upscaling with unpooling layers that resemblance the filter and muxing structure of classic upscalers. We showed how to formally interpret a convolutional network as a space--variant linear system when activation functions are fixed. And we introduced algorithms to extract this linear structure from a convolutional network. Our network structure has been succesfully applied for image upscaling, using the same model to optimize an image quality metric (SSIM) or to generate upscaled images with artificial details that can look as real as original content.

\bibliographystyle{IEEEbib}
\bibliography{refs}

\end{document}